\documentclass[10pt]{article}
\usepackage[utf8]{inputenc} % allow utf-8 input
\usepackage[T1]{fontenc}    % use 8-bit T1 fonts
\usepackage{hyperref}       % hyperlinks
\usepackage{url}            % simple URL typesetting
\usepackage{booktabs}       % professional-quality tables
\usepackage{amsfonts}       % blackboard math symbols
\usepackage{nicefrac}       % compact symbols for 1/2, etc.
\usepackage{microtype}      % microtypography

\usepackage{algorithmicx,algorithm}

\usepackage{amsmath, amsthm, amssymb, multirow, paralist}
\usepackage[numbers]{natbib}

\usepackage[noend]{algpseudocode}
\algblock{Input}{EndInput}
\algnotext{EndInput}
\algblock{Output}{EndOutput}
\algnotext{EndOutput}
\newcommand{\Desc}[2]{\State \makebox[3em][l]{#1}#2}

\usepackage{graphicx}

\usepackage{comment}
\usepackage{color}
\usepackage{dsfont}

\usepackage{wrapfig}

\usepackage[body={6.5in,8in}]{geometry}

\def \R {\mathbb{R}}

\title{Revisiting Sample Selection Approach to Positive-Unlabeled Learning:
Turning Unlabeled Data into \\Positive rather than Negative}

\date{}

\author{
  Miao Xu$^1$, Bingcong Li, Gang Niu$^{1}$, Bo Han$^{1,2}$, Masashi Sugiyama$^{1,3}$\\
  $^1$RIKEN Center for Advanced Intelligence Project, Tokyo, Japan\\
  $^2$Center for Artificial Intelligence, University of Technology Sydney, Sydney, Australia\\
  $^3$University of Tokyo, Tokyo, Japan\\
  \texttt{\{miao.xu, gang.niu, bo.han\}@riken.jp}, \\
  \texttt{bingcon.li@gmail.com},\\
  \texttt{sugi@k.u-tokyo.ac.jp}
}

\begin{document}

\maketitle

\begin{abstract}
In the early history of positive-unlabeled (PU) learning, the \emph{sample selection approach}, which heuristically selects negative (N) data from U data, was explored extensively. However, this approach was later dominated by the \emph{importance reweighting approach}, which carefully treats all U data as N data. May there be a new sample selection method that can outperform the latest importance reweighting method in the deep learning age? This paper is devoted to answering this question affirmatively---we propose to label large-loss U data as P, based on the memorization properties of deep networks. Since P data
selected in such a way are biased, we develop a novel learning objective that can handle such biased P data properly. Experiments
confirm the superiority of the proposed method.
\end{abstract}

%\section{Further Discussion (for internal use only)}

\section{Introduction}
In machine learning, there are situations where only positive labeled data can be collected, together with a large amount of unlabeled data. Learning with such data is called positive-unlabeled (PU) learning~\cite{DBLP:conf/alt/Denis98}. Since PU learning can be seen as learning with clean P data and noisy negative (N) data, it is also related to learning with noisy labels~\cite{DBLP:journals/ml/AngluinL87,DBLP:conf/colt/ScottBH13} and UU learning~\cite{DBLP:conf/iclr/LuNMS19}, while in the latter two learning paradigms \emph{no} clean P data is available.

Depending on different data generation processes, problem setting of PU learning can be divided into two categories: censoring PU learning~\cite{DBLP:conf/kdd/ElkanN08} where U data is collected first and then some of P data in the U data is labeled, and case-control PU learning~\cite{doi:10.1111/j.1541-0420.2008.01116.x} where P and U data are collected separately.  Case-control PU learning is slightly more general than censoring PU learning~\cite{DBLP:conf/icml/MenonROW15} and thus nowadays there are many efforts devoted to solving it~\cite{DBLP:conf/nips/PlessisNS14,DBLP:conf/icml/PlessisNS15,DBLP:conf/icml/SakaiPNS17,DBLP:conf/nips/KiryoNPS17}. In this paper, we also focus on the case-control PU learning. 

Many PU learning methods have been proposed and they can be generally divided into generative and discriminative methods. Discriminative methods are the majority while there is few generative algorithm except~\cite{DBLP:conf/ijcai/HouCLZ18} which relies on the restrictive cluster assumption~\cite{DBLP:conf/nips/ChapelleWS02}. In the early development of discriminative methods, the \emph{sample selection approach} was proposed, which selects N data from U data and performs ordinary PN learning afterwards. Various heuristics were designed then for selecting reliable N data, including~\cite{DBLP:conf/ijcai/LiL03},~\cite{DBLP:conf/icml/LiuLYL02} and~\cite{DBLP:conf/kdd/YuHC02}. However, such a trend was soon braked by~\cite{DBLP:conf/icml/LeeL03} and~\cite{DBLP:conf/icdm/LiuDLLY03}, which proposed the \emph{importance reweighting approach} treating U data directly as N data but with reduced importance weights. Such an approach was compared extensively to various sample selection strategies~\cite{DBLP:conf/icdm/LiuDLLY03}, and experimental results showed that the former is significantly better. The belief that importance reweighting approach is superior to the sample selection approach was further strengthened~\cite{DBLP:conf/kdd/ElkanN08}. From then on,  the importance reweighting methods of PU learning continue to be developed. Representative methods are based on unbiased risk estimation~\cite{DBLP:conf/nips/PlessisNS14,DBLP:conf/icml/PlessisNS15}, which rewrites the classification risk to an equivalent form depending only on PU data. Among them, the \emph{unbiased PU} (uPU)~\cite{DBLP:conf/icml/PlessisNS15} achieved relatively better empirical results and are accompanied with strong theoretical guarantees~\cite{DBLP:conf/nips/NiuPSMS16}.

In the original uPU paper~\cite{DBLP:conf/icml/PlessisNS15}, simple models with only a small number of parameters were used. However, if complex models such as deep networks are used, serious overfitting happens while a breakthrough for this problem is~\cite{DBLP:conf/nips/KiryoNPS17}. It realized that the overfitting of uPU is due to the rewrite of the risk to include a subtraction calculation. When minimizing the empirical risk during training, beacause of the strong capacity of complex models the risk will reach a minimum long away below zero, while it should be no less than zero. To deal with this problem,~\cite{DBLP:conf/nips/KiryoNPS17} proposed the \emph{non-negative PU} (nnPU) method, which performs regular gradient descent if the training loss is above zero, and gradient ascent otherwise. Such a non-negative idea has been generalized to more general scenarios recently \cite{DBLP:journals/corr/abs-1809-11008}.%}. 

Note that nnPU is still an importance reweighting method. While aforementioned belief dated back in 2008 was based on simple models, it may not hold on complex models nowadays. Since complex models are used widely~\cite{Goodfellow-et-al-2016}, we are wondering whether there is a possibility for the sample selection approach in this deep learning age.

In this paper, we answer this question affirmatively by:
%\vspace{-3mm}
\begin{itemize}%[leftmargin=*]
\item 
What to select? We were inspired by the theoretical result that using simple models, P data will benefit PU learning more than PN learning ~\cite{DBLP:conf/nips/NiuPSMS16}. We empirically verify the importance of P data for PU learning using complex models and propose to select P.
\item 
How to select? Recent work about the memorization properties of deep networks~\cite{DBLP:conf/icml/ArpitJBKBKMFCBL17} inspired us. The memorization properties were used in~\cite{DBLP:conf/icml/JiangZLLF18} and~\cite{DBLP:conf/nips/HanYYNXHTS18} treating small-loss data as clean data. According to them, large-loss data would be P data in PU learning. Such an assumption is confirmed empirically. We also discuss the choices of surrogate loss functions to make the selected P cleaner. 
%\vspace{-0.05in}
\item 
How to use the selected data? We show that the selected data is usually biased and using them directly causes serious underestimation of the empirical risk. We then carefully design a new learning objective for the selected biased data to be used as P data. 
\end{itemize}
%\vspace{-3mm}
We call our method \emph{adaptively augmented PU} (aaPU).  aaPU automatically identifies P data from a set of U data during the training process. In each epoch, the P data is selected and used for further training. Experimental results show that aaPU is superior. 

Our contribution lies in three aspects.
%\vspace{-3mm}
\begin{itemize}%[leftmargin=*]
\item 
Compared to existing sample selection methods in PU learning~\cite{DBLP:conf/icdm/LiuDLLY03,DBLP:conf/icml/LiuLYL02,DBLP:conf/kdd/YuHC02,DBLP:conf/ijcai/LiL03}, our proposal is end-to-end and selects \emph{positive} instead of \emph{negative} data; more importantly, classical methods cannot work with complex models, because they select samples by training a PN classifier treating U data naively as N data. As we will show in Figure~\ref{fig:loss-distribution-upu} when complex models are used, all U data will be classified to be N after training even if we do optimal reweighting based on unbiased risk estimators. 
\item 
Our proposal adds the biased selected data \emph{permanently}, which is more difficult than adding them \emph{temporarily} in learning with noisy labels' works~\cite{DBLP:conf/icml/JiangZLLF18,DBLP:conf/nips/HanYYNXHTS18} exploiting the same ``large-small-loss'' trick.
\item 
One important direction in semi-supervised learning~\cite{DBLP:reference/ml/Zhu17} is self-training~\cite{DBLP:journals/tit/Scudder65a}. Such a learning approach requires labeled data from \emph{all} classes to begin with while in many weakly supervised problems~\cite{doi:10.1093/nsr/nwx106}, we only have labeled data from \emph{some} classes. To the best of our knowledge, this is the first work to explore this direction.
\end{itemize}
%\vspace{-2mm}
%\paragraph{Organization}
The rest of the paper is organized as follows. In Sec.~\ref{sec:formulation} we give the formulation. In Sec.~\ref{sec:proposal}, we answer what to select, how to select and how to use the selected data, ended with the algorithm framework. We give the experimental results in Sec.~\ref{sec:experiments} and conclude in Sec.~\ref{sec:conclusion} with future work.

\section{Formulation and Review}\label{sec:formulation}

In this section, we introduce the formulation, as well as the risk used in PU learning. 

Let $X\in\R^d$ be the input random variable and denote the output random variable by $Y\in \{1,-1\}$. In PU learning, the training set $D$ is composed of two parts, positive data $\mathcal{X}_\mathrm{p}$ and unlabeled data $\mathcal{X}_\mathrm{u}$. $\mathcal{X}_\mathrm{p}$ contains $n_\mathrm{p}$ instances sampled from $P(x|Y=1)$. $\mathcal{X}_\mathrm{u}$ contains $n_\mathrm{u}$ instances sampled from $P(x)$.  Denote by $\pi$ the class-prior probability, i.e., $\pi=P(Y=1)$. Same as~\cite{DBLP:conf/nips/KiryoNPS17}, we assume that $\pi$ is known throughout the paper. In practice, it can be estimated from the given PU data~\cite{DBLP:conf/icml/RamaswamyST16}. 

let the parameters of the classification model be $\theta$, and $g(x;\theta)$ be the decision function. To learn a classifier, we need to minimize the risk 
\begin{eqnarray}\label{eq:general-risk}
R(g)=\mathbb{E}_{(X,Y)\sim p(x,y)}[\ell(Yg(X;\theta))].
\end{eqnarray} 
where $\ell(\cdot)$ is any trainable surrogate loss of the zero-one loss. Among the loss functions in~\cite{DBLP:conf/nips/KiryoNPS17}, the most popular one is the \emph{sigmoid loss}, which is defined as
\begin{eqnarray}\label{eq:sigloss}
\ell_\mathrm{sigmoid}(ty)= 1/(1+\exp(ty)).
\end{eqnarray}
There is also other surrogate loss functions, such as \emph{logistic loss} defined as
\begin{eqnarray}\label{eq:logisticloss}
\ell_\mathrm{logistic}(ty)=\ln(1+\exp(-ty)).
\end{eqnarray}
We will discuss the choice of surrogate loss in Sec.~\ref{sec:how-to-select}. 

%With the surrogate loss $\ell_\mathrm{s}(t,y)$, generally the learning objective of $g$ is given by, 
%\begin{eqnarray}\label{eq:general-surrogate-risk}
%R(g)=\mathbb{E}_{(X,Y)\sim p(x,y)}[\ell_\mathrm{s}(g(X;\theta),Y)].
%\end{eqnarray} 
For PN learning, the risk can be written as
\begin{eqnarray}\label{eq:pn-loss}
R_\mathrm{pn}(g)=\pi\mathbb{E}_{X\sim p(x|Y=1)}[\ell(g(X;\theta))]+\;\;\;\;\;\;\;\;\;\;\;\;\\\nonumber
(1-\pi)\mathbb{E}_{X\sim p(x|Y=-1)}[\ell(-g(X;\theta))].
\end{eqnarray}
and its empirical estimation would be
\begin{eqnarray}\label{eq:empirical-pn-loss}
\widehat R_\mathrm{pn}= \frac{\pi}{n_\mathrm{p}}\sum_{i=1}^{n_\mathrm{p}} \ell(g(x_i;\theta))+\frac{1-\pi}{n_\mathrm{n}}\sum_{i=1}^{n_\mathrm{n}} \ell(-g(x_i;\theta))
\end{eqnarray}
where $n_\mathrm{n}$ is the number of N data.

In PU learning, because we do not have negative data, with
\begin{eqnarray*}
p(x)=\pi p(x|Y=1)+(1-\pi) p(x|Y=-1),
\end{eqnarray*}
we can replace the negative loss part in Eq.~(\ref{eq:pn-loss}) by
\begin{align}\label{eq:negative-loss}
&(1-\pi)\mathbb{E}_{X\sim p(x|Y=-1)}[\ell(-g(X;\theta))]=\\\nonumber
&\mathbb{E}_{X\sim p(x)}[\ell(-g(X;\theta))] %\\\nonumber
-\pi\mathbb{E}_{X\sim p(x|Y=1)}[\ell(-g(X;\theta))].
\end{align}

To simplify our description in the following, we denote
\begin{eqnarray*}
%R^+ &=&\mathbb{E}_{X\sim p(x|Y=1)}[\ell(Yg(X;\theta))],\\
%R^- &=&\mathbb{E}_{X\sim p(x|Y=1)}[\ell(Yg(X;\theta))],\\
R_\mathrm{p}^+ &=&\mathbb{E}_{X\sim p(x|Y=1)}[\ell(g(X;\theta))],\\
R_\mathrm{p}^- &=&\mathbb{E}_{X\sim p(x|Y=1)}[\ell(-g(X;\theta))],\\
R_\mathrm{n}&=&\mathbb{E}_{X\sim p(x|Y=-1)}[\ell(-g(X;\theta))], \\
R_\mathrm{u}&=&\mathbb{E}_{X\sim p(x)}[\ell(-g(X;\theta))].
\end{eqnarray*}
In this way, the expected risk for PU learning is written as
\begin{eqnarray}\label{eq:expect-pu-risk}
R_\mathrm{pu}=\pi R_\mathrm{p}^++(R_\mathrm{u}-\pi R_\mathrm{p}^-).
\end{eqnarray}
Given the empirical estimation of risk in Eq.~(\ref{eq:expect-pu-risk}), which is
\begin{eqnarray}\label{eq:empirical-pu-risk}
&&\widehat R_\mathrm{pu}= \frac{\pi}{n_\mathrm{p}}\sum_{i=1}^{n_\mathrm{p}} \ell(g(x_i;\theta))+\\\nonumber
&&\left(\frac{1}{n_\mathrm{u}}\sum_{i=1}^{n_\mathrm{u}} \ell(-g(x_i;\theta))-\frac{\pi}{n_\mathrm{p}}\sum_{i=1}^{n_\mathrm{p}} \ell(-g(x_i;\theta))\right),
\end{eqnarray}
and minimizing it, we can have the uPU method~\cite{DBLP:conf/nips/PlessisNS14}. 

For nnPU~\cite{DBLP:conf/nips/KiryoNPS17}, it realized that in uPU by minimizing Eq.~(\ref{eq:empirical-pu-risk}) using flexible enough deep networks, the second line of Eq.~(\ref{eq:empirical-pu-risk}), which should have stayed positive goes negative. Thus they proposed to optimizing a non-negative version of Eq.~(\ref{eq:empirical-pu-risk}), which is 
\begin{align}\label{eq:loss-nnpu}
\widehat R_\mathrm{nnPU}&= \frac{\pi}{n_\mathrm{p}}\sum_{i=1}^{n_\mathrm{p}} \ell(g(x_i;\theta))+\max\bigg(0,\\\nonumber &
\frac{1}{n_\mathrm{u}}\sum_{i=1}^{n_\mathrm{u}} \ell(-g(x_i;\theta))-\frac{\pi}{n_\mathrm{p}} \sum_{i=1}^{n_\mathrm{p}} \ell(-g(x_i;\theta))\bigg), 
\end{align}
%Although the detailed implementation is a bit different, in that they used some hyper parameters to add into flexibility.
In the practical implementation, nnPU checked whether the loss $\widehat R_\mathrm{pu}$ is larger than zero. If it is larger, gradient descent is performed. Otherwise, nnPU performs \emph{gradient ascent} to correct the overfitting effect.

\section{Our Method}\label{sec:proposal}

In this section, we discuss what to select, how to select and how to use the selected data. Finally, we summarize the proposed algorithm.
\subsection{What to Select}
Considering what to select, we are motivated by a conclusion in~\cite{DBLP:conf/nips/NiuPSMS16}. Let the minimizer of $R_\mathrm{pn}(g)$ in Eq.~(\ref{eq:empirical-pn-loss}) be $\widehat g_\mathrm{pn}$ and the minimizer of $R_\mathrm{pu}(g)$ in Eq.~(\ref{eq:empirical-pu-risk}) be $\widehat g_\mathrm{pu}$. Additionally, assume $g^*$ optimizes $R(g)$ in Eq.~(\ref{eq:general-risk}). In~\cite{DBLP:conf/nips/NiuPSMS16}, Corollary 5 states that with conditions on the functional class and probability at least $1-\delta$, 
\begin{eqnarray*}
R(\widehat g_\mathrm{pn})-R(g^*)&\le& f(\delta)\cdot\{\pi/\sqrt{n_\mathrm{p}}+(1-\pi)/\sqrt{n_\mathrm{n}}\},\\
R(\widehat g_\mathrm{pu})-R(g^*)&\le& f(\delta)\cdot\{2\pi/\sqrt{n_\mathrm{p}}+1/\sqrt{n_\mathrm{u}}\}.
\end{eqnarray*} 

\begin{figure}[!t]
\centering
\begin{minipage}[h]{2.6in}
\centering
\includegraphics[width= 2.6in]{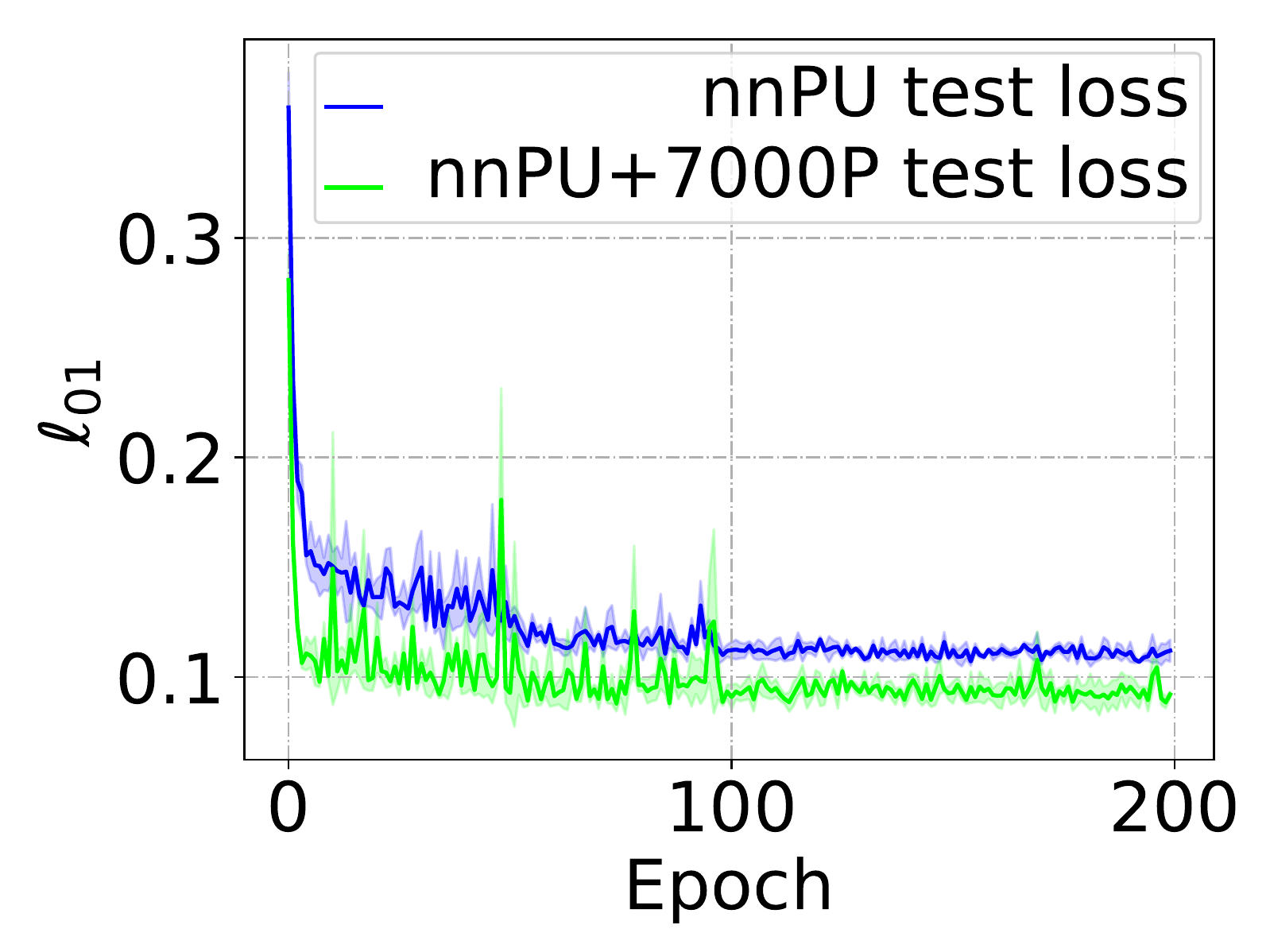}\\
%\mbox{(a)}
\end{minipage}
\caption{\small  The effect of adding positive data for PU learning on test loss of the CIFAR-10 dataset~\cite{Krizhevsky09learningmultiple}. We used nnPU~\cite{DBLP:conf/nips/KiryoNPS17} as a classifier and randomly sampled $7,000$ additional P data in the nnPU+P method. The shadow shows the variance of each method}\label{fig:add-p-motivation}
\end{figure}

An interesting conclusion we can derive from the theoretical result above is that, adding more P data besides the original P data will benefit PU learning more than PN learning, since the weights of the empirical average of P data is $2\pi$ in PU learning, which is twice of that in PN learning~\cite{DBLP:conf/nips/NiuPSMS16}. Although the theoretical results are based on linear-in-parameter model due to the difficulty of deriving the Rademacher complexity~\cite{DBLP:journals/jmlr/BartlettM02} on complex deep networks, we were motivated by such a conclusion and performed some empirical studies. In Figure~\ref{fig:add-p-motivation}, we show the zero-one test loss on the CIFAR-10 data~\cite{Krizhevsky09learningmultiple} comparing two methods: nnPU~\cite{DBLP:conf/nips/KiryoNPS17}, nnPU with $7,000$ more positive data (nnPU+P) using a deep neural network as a classification model. We used the same network structure and setting as that in Sec.~\ref{sec:exp-real}. From the results, we can see that for deep networks, adding positive data can benefit PU learning. Thus in the following, we will consider selecting P data from U data, instead of N data.

\subsection{How to Select}\label{sec:how-to-select}
\begin{figure}[!t]
\centering
\begin{minipage}[h]{2.4in}
\centering
\includegraphics[width= 2.4in]{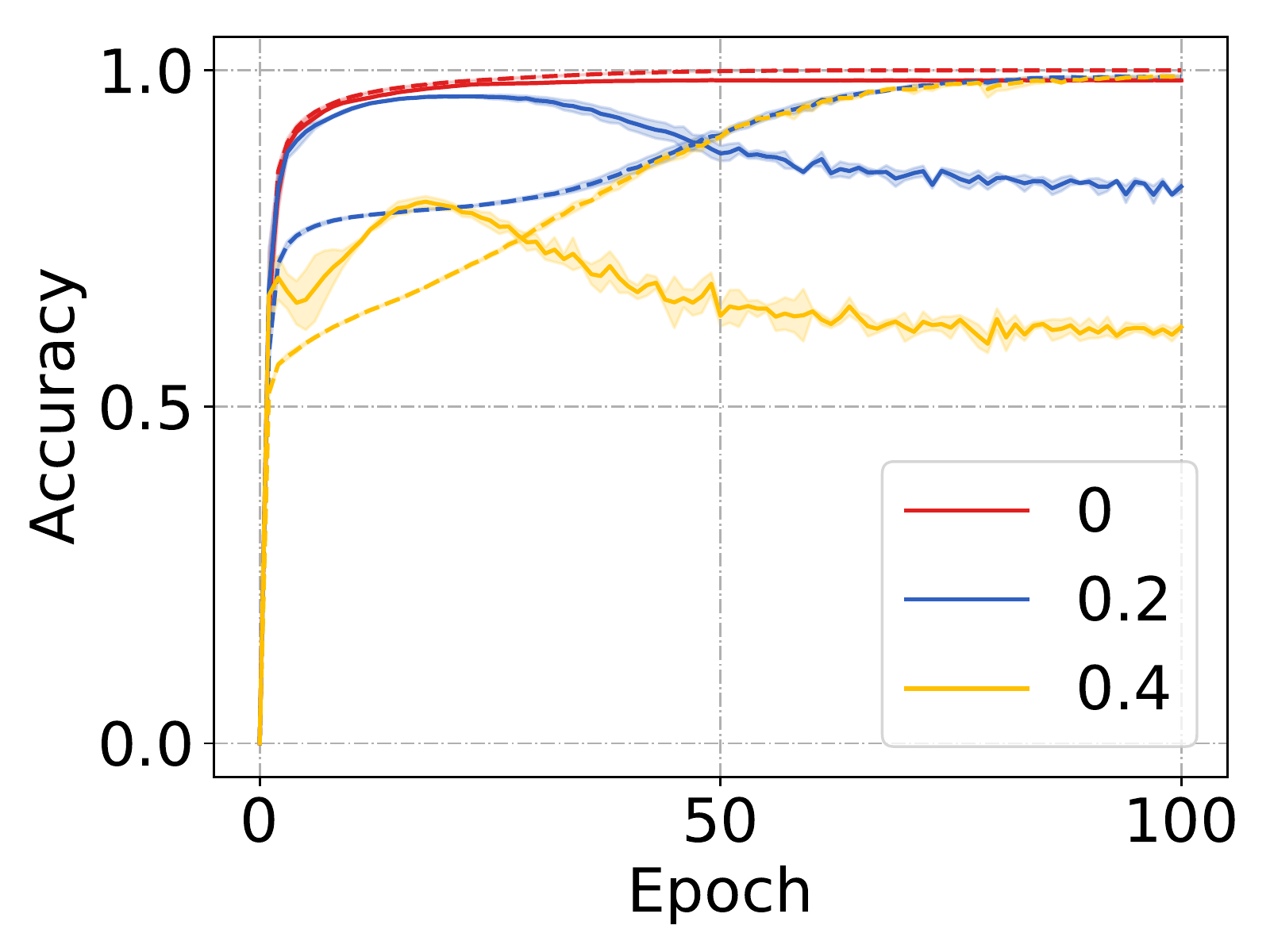}
%\mbox{(a)}
\end{minipage}
\caption{\small  Accuracy on training data (dotted) and validation data (solid), where $0$ means no noise, and $0.2$ ($0.4$) means $20\%$ ($40\%$) random label noise. This is re-implementation of part of the results in Figure 7~\cite{DBLP:conf/icml/ArpitJBKBKMFCBL17}. }\label{fig:memorization}
\end{figure}

Regarding how to select, we were motivated by recent work about the memorization properties of deep neural networks~\cite{DBLP:conf/icml/ArpitJBKBKMFCBL17}, showing that during the process of stochastic gradient descent, deep neural networks tend not to memorize all data at the same time, but memorize frequent patterns first and later irregular patterns.  Such an effect is verified in~\cite{DBLP:conf/icml/ArpitJBKBKMFCBL17} through empirical studies on learning with noisy labels problem~\cite{DBLP:conf/nips/NatarajanDRT13,DBLP:conf/icml/MenonROW15,DBLP:journals/pami/LiuT16,DBLP:conf/icml/MaWHZEXWB18}. We re-implmented part of the result in Figure~\ref{fig:memorization}. From this, we can see that the validation accuracies increase in the first few epochs on both clean and noise data fitting the frequent patterns. However, on noisy data they suddenly drop after the first few epochs, showing that they begin to fit noise, i.e., irregular patterns. 

Researchers in the community of learning with noisy labels have been using such memorization properties to differentiate noisy and non-noisy data by large and small loss respectively~\cite{DBLP:conf/icml/JiangZLLF18,DBLP:conf/nips/HanYYNXHTS18}. Such works give us good intuition on how to identify P data from U data, as U data in PU learning can be seen as noisy N data. However, we do not know whether such a ``large-small-loss-trick'' still works for PU learning. We expect that PU learning algorithms based on deep networks will remember N data first, and later P data. 

\begin{figure*}[!t]
\centering
\begin{minipage}[h]{2.0in}
\centering
\includegraphics[width= 2.0in]{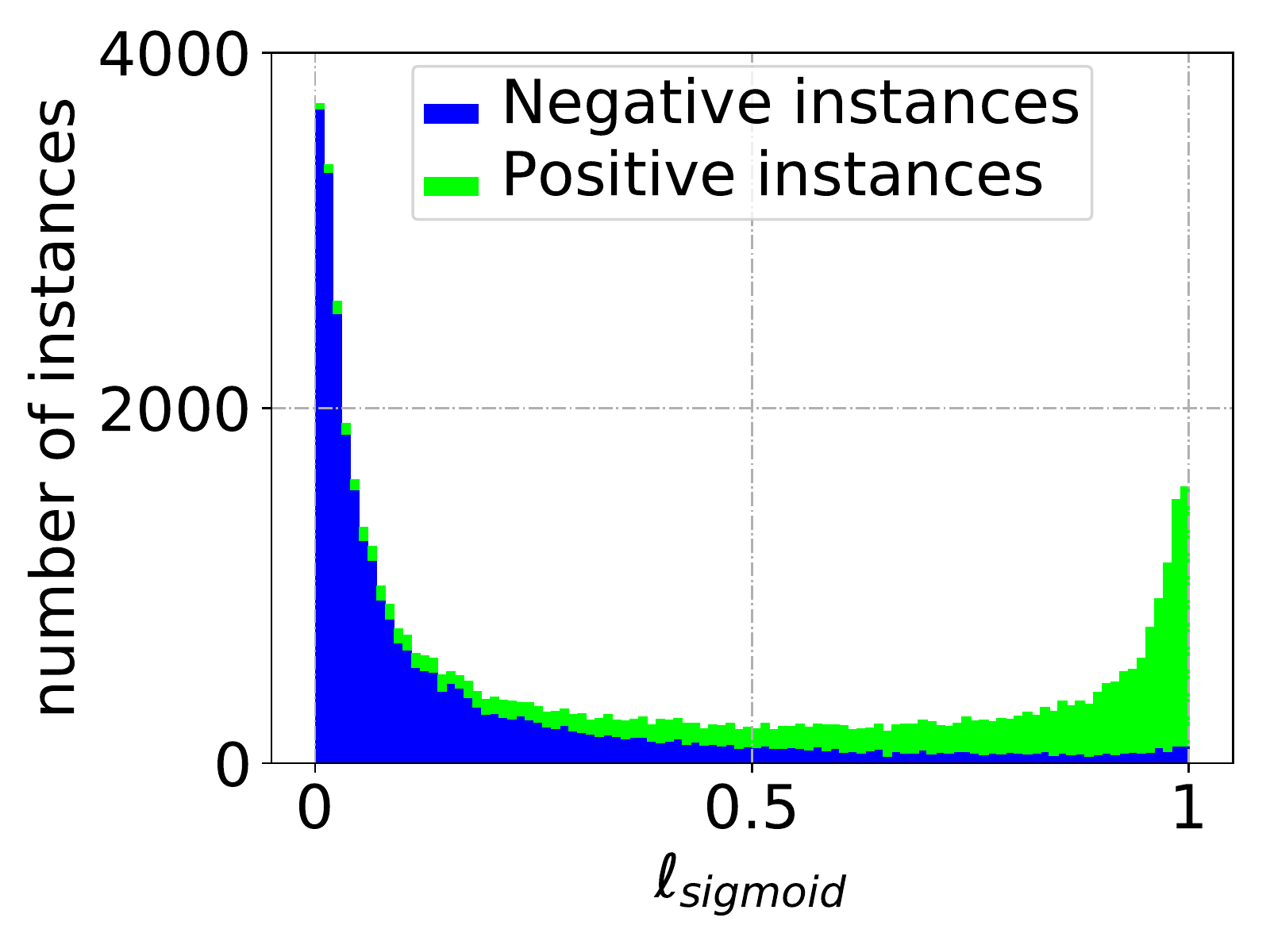}\\
\mbox{(a)} \small{Epoch $10$}
\end{minipage}
\begin{minipage}[h]{2.0in}
\centering
\includegraphics[width= 2.0in]{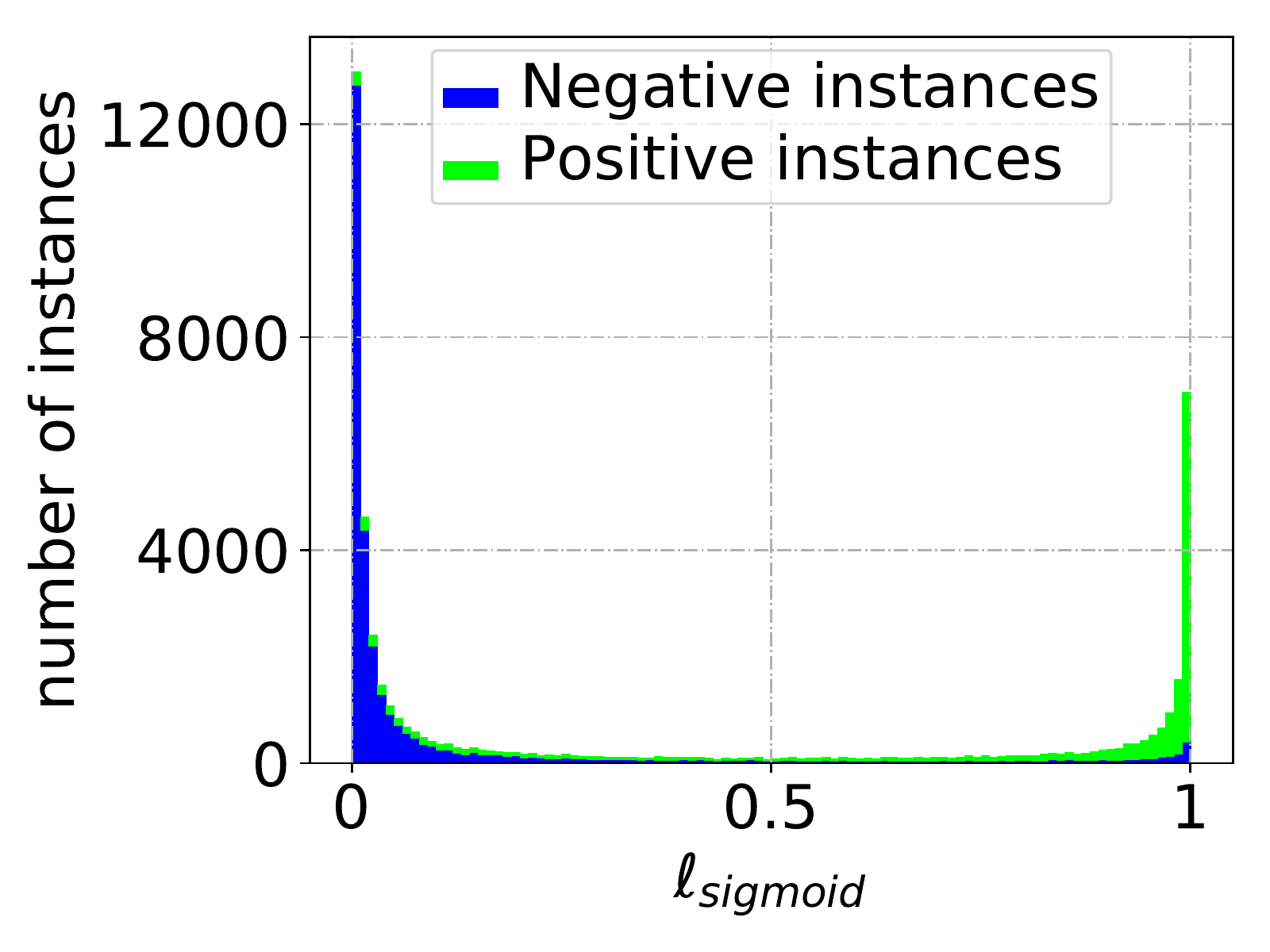}\\
\mbox{(b)}  \small{Epoch $50$}
\end{minipage}
%\centering
%\begin{minipage}[h]{2.0in}
%\centering
%\includegraphics[width= 2.0in]{img/fig1/hist_100}\\
%\mbox{(c)}  Epoch $100$
%\end{minipage}
\begin{minipage}[h]{2.0in}
\centering
\includegraphics[width= 2.0in]{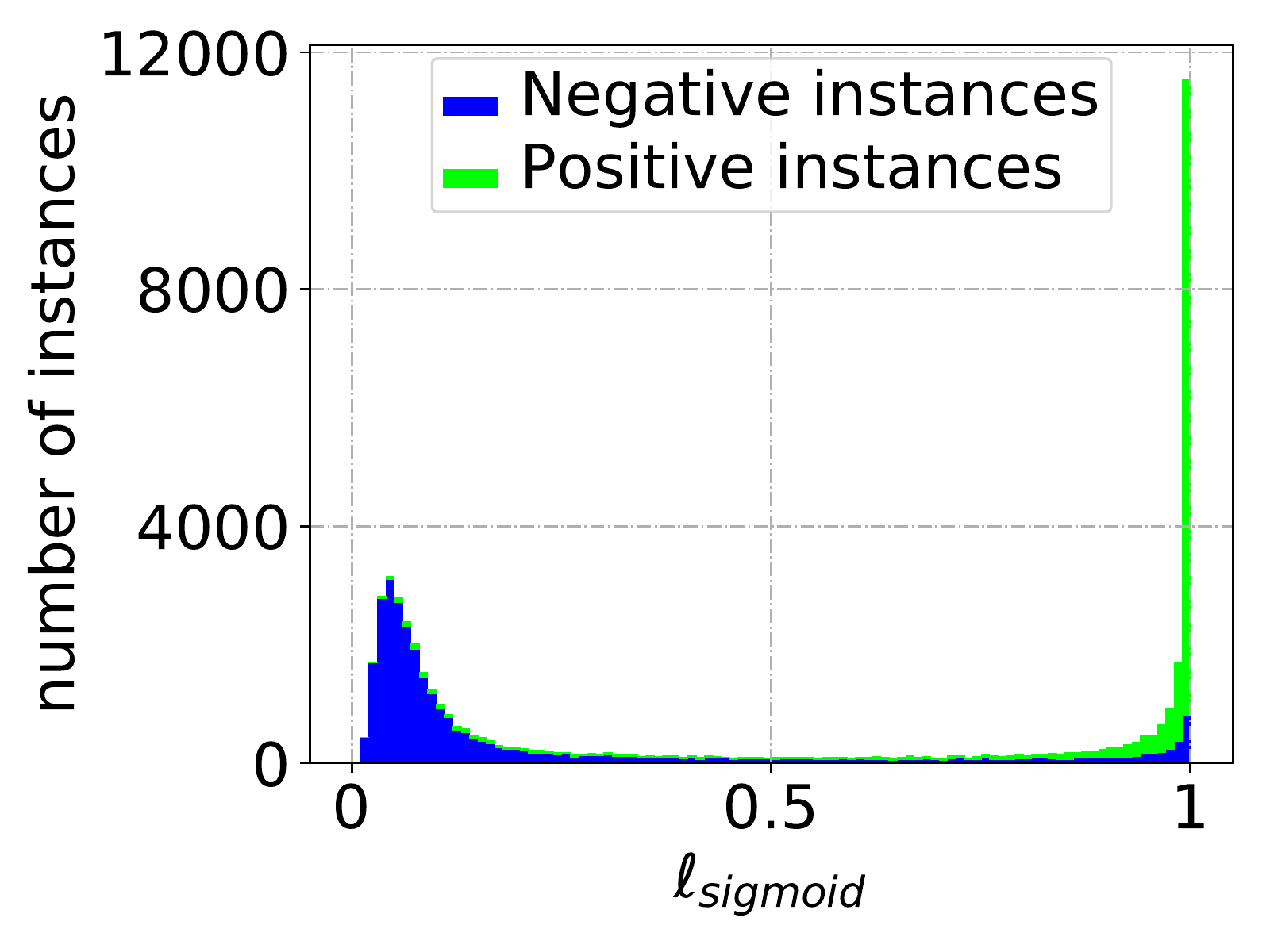}\\
\mbox{(c)} \small{Epoch $200$}
\end{minipage}
\caption{\small The nnPU algorithm's sigmoid loss histogram of U data on CIFAR-10. We feed forward all U data into the neural network and calculate their sigmoid loss when treating them as negative. We then divide the range $[0,1]$ equally into $100$ bins and calculate the number of instances falling into each bin. (a) Sigmoid loss histogram on epoch $10$; (b) sigmoid loss histogram on epoch $50$; (c) sigmoid loss histogram on epoch $200$.}\label{fig:loss-distribution-nnpu}
\end{figure*}

\begin{figure*}[!t]
\centering
\begin{minipage}[h]{2.0in}
\centering
\includegraphics[width= 2.0in]{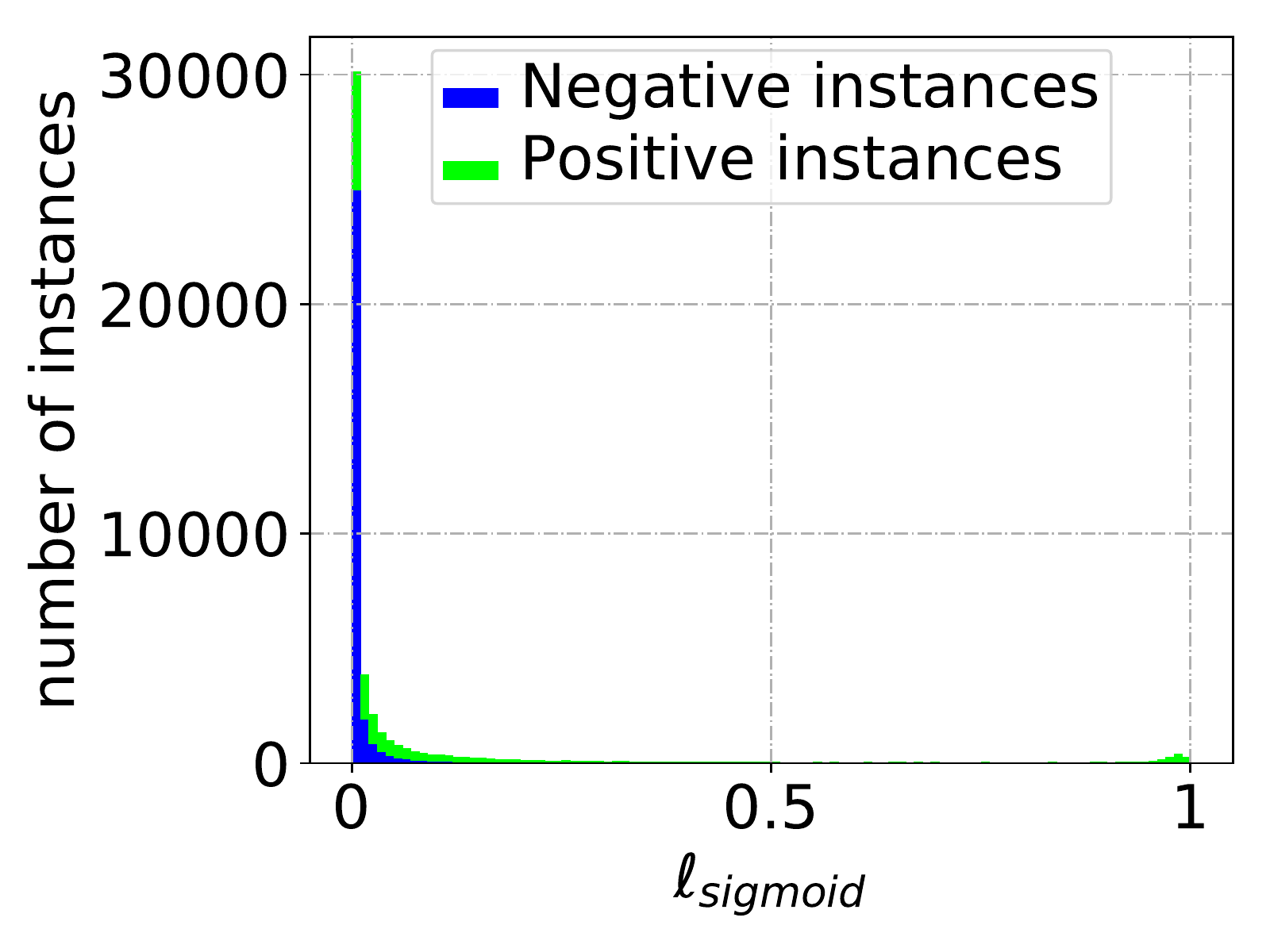}\\
\mbox{(a)} \small {Epoch $10$}
\end{minipage}
\begin{minipage}[h]{2.0in}
\centering
\includegraphics[width= 2.0in]{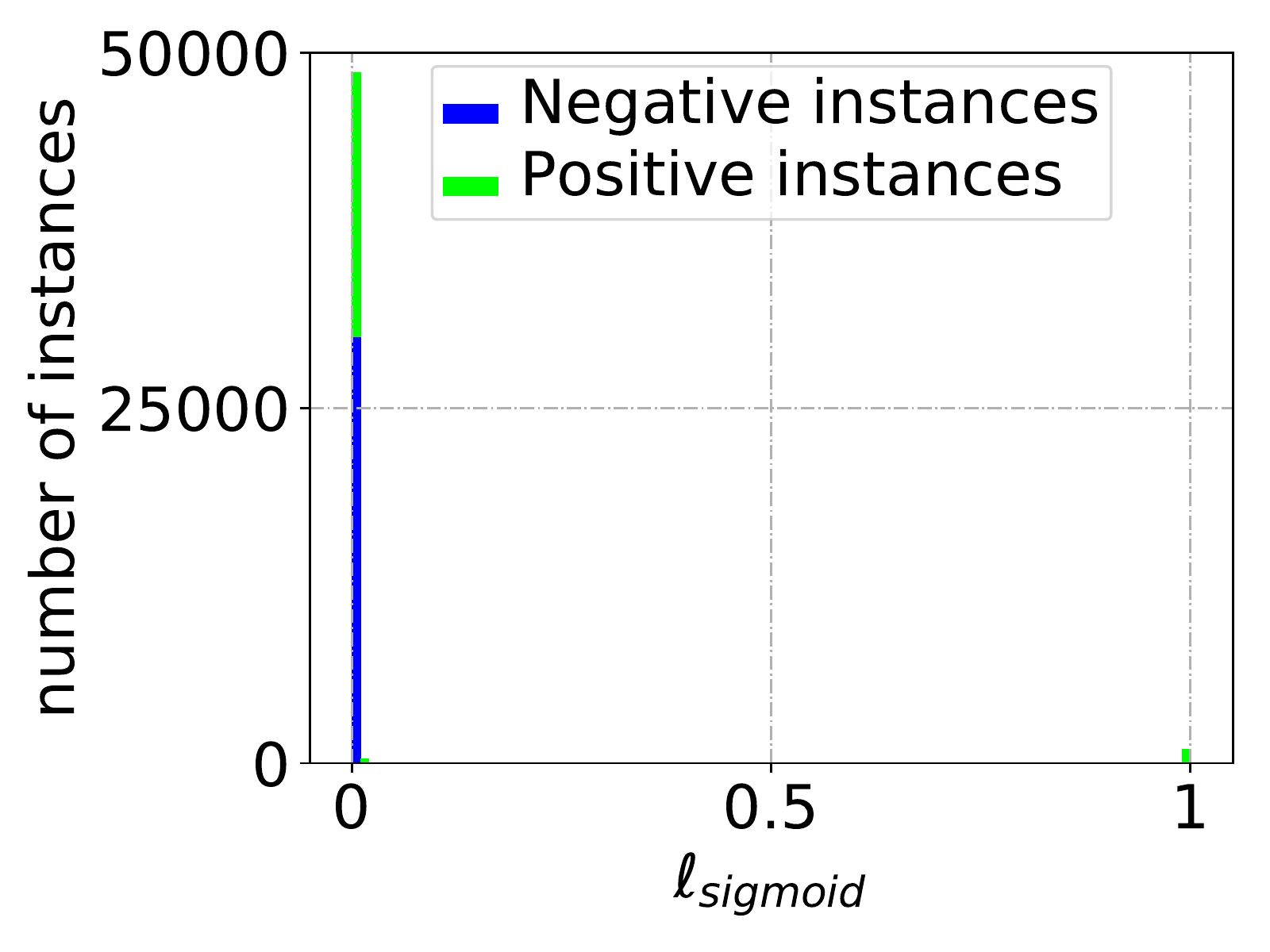}\\
\mbox{(b)}  \small{Epoch $50$}
\end{minipage}
%\centering
%\begin{minipage}[h]{2.0in}
%\centering
%\includegraphics[width= 2.0in]{img/fig2/hist_100}\\
%\mbox{(c)}  Epoch $100$
%\end{minipage}
\begin{minipage}[h]{2.0in}
\centering
\includegraphics[width= 2.0in]{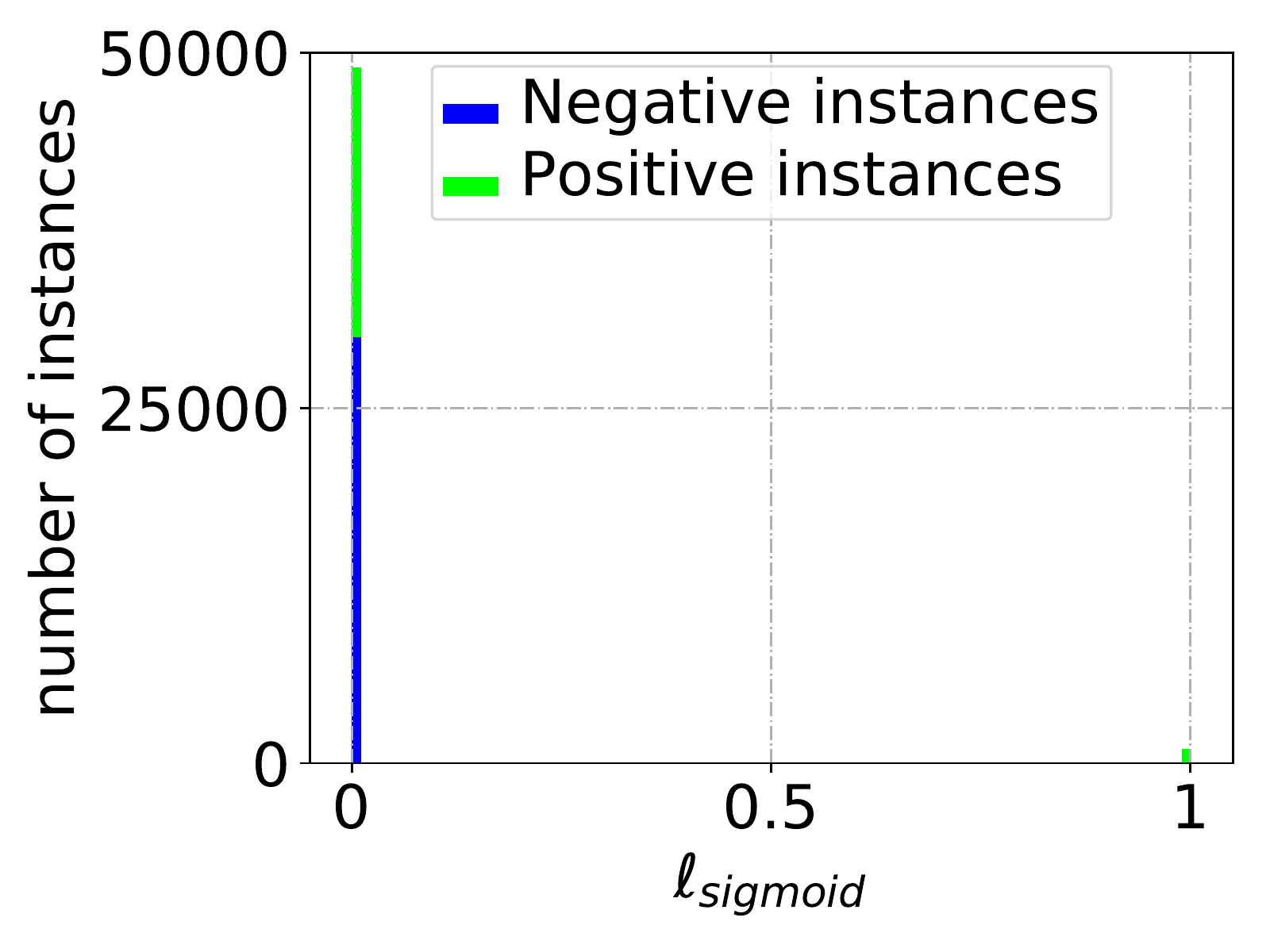}\\
\mbox{(c)} \small{Epoch $200$}
\end{minipage}
\caption{\small The uPU algorithm's sigmoid loss histogram of U data on CIFAR-10. We feed forward all U data into the neural network and calculate their sigmoid loss when treating them as negative. We then divide the range $[0,1]$ equally into $100$ bins and calculate the number of instances falling into each bin. (a) Sigmoid loss histogram on epoch $10$; (b) sigmoid loss histogram on epoch $50$; (c) sigmoid loss histogram on epoch $200$.}\label{fig:loss-distribution-upu}
\end{figure*}

To confirm our expectation, empirically,  after finishing a particular epoch, we fed forward all U data into the current neural network. Since these data do not have labels, we calculate their loss by assuming their labels are negative.  We then plotted the histogram of these loss to show their distribution. Specially, we divided the range $[0,1]$ of these loss equally into $100$ bins and counted the number of loss falling into each bin. The histogram on epochs $10$, $50$, and $200$ for nnPU are shown in Figure~\ref{fig:loss-distribution-nnpu} where the true N data is marked \emph{blue} and true P data is marked \emph{green}. Obviously most green P data have large loss. Besides this phenomenon, we can see the following tendencies, 
%\vspace{-0.1in}
%\vspace{-3mm}
\begin{itemize}%[leftmargin=*]
\item The ``large-small loss'' trick also works in PU learning. We can see in Figure~\ref{fig:loss-distribution-nnpu} that as the training process evolves from epoch $10$ to epoch $50$, the majority of large-loss data are P data. Thus such data can be identified by splitting U data according to some threshold on the loss. However, when the training process continues until epoch $200$, large-loss data selected as positive are containing more N data, i.e., more noise. This shows that we should select positive data at appropriate epochs and eventually we could reduce the number of data selected.
\item For nnPU, although we can get a reasonable amount of positive data, large-loss data are not purely positive as shown in Figure~\ref{fig:loss-distribution-nnpu}. By selecting such positive data, we may get some noise. During the training process the effect of noise can be accumulated and eventually deteriorate the performance of the algorithm. 
\end{itemize}
%\vspace{-3mm}
Our next task is to consider how to select P data as \emph{pure} as possible. First, we note that nnPU performs gradient \emph{ascent} during training to prevent overfitting which tends to classify some N data as P. In this way, we try the uPU method with complex models. If uPU can give us a reasonable number of pure P data, then, even if uPU overfits, we can still use it as a sample selection approach. The histogram of loss of U data on epochs $10$, $50$, and $200$ for uPU are shown in Figure~\ref{fig:loss-distribution-upu}. From Figure~\ref{fig:loss-distribution-upu}, we can see that although the large-loss part are purely ``positive'', these data are the overlapping of P and U data, thus they are useless. In the following, we will do sample selection with nnPU.
%Although the nnPU cannot give perfect positive data, the performance of aaPU based on it can still be empirically verified as in Section~\ref{sec:experiments}. There are also some parameters in nnPU to control the difference between nnPU and uPU. If needed, such parameters can be further tuned to get a better performance.   

We consider another way to get pure P data---choosing a better surrogate loss function. Previous works such as nnPU~\cite{DBLP:conf/nips/KiryoNPS17} used the sigmoid loss in Eq.~(\ref{eq:sigloss}) as the surrogate loss. Although such surrogate loss function is useful in previous works, it has a frawback in our context, i.e., its values are always within the range of $[0,1]$. Note that in our work, we need to use the ``large-small loss'' trick to identify large-loss data in the set of U data as P. However, as the sigmoid loss is bounded from above, the loss of positive data are also upper-bounded. Thus even if these positive data have a potential to reach larger loss, they are upper limited. Among all the surrogate loss functions listed in~\cite{DBLP:conf/nips/KiryoNPS17}, we found that the logistic loss in Eq.~(\ref{eq:logisticloss}) is not upper bounded. It also shares some of other merits of the sigmoid loss, such as being Lipschitz continuous and differentiable everywhere. We thus use the logistic loss instead of the sigmoid loss.  

\begin{figure*}[!t]
\centering
\begin{minipage}[h]{2.0in}
\centering
\includegraphics[width= 2.0in]{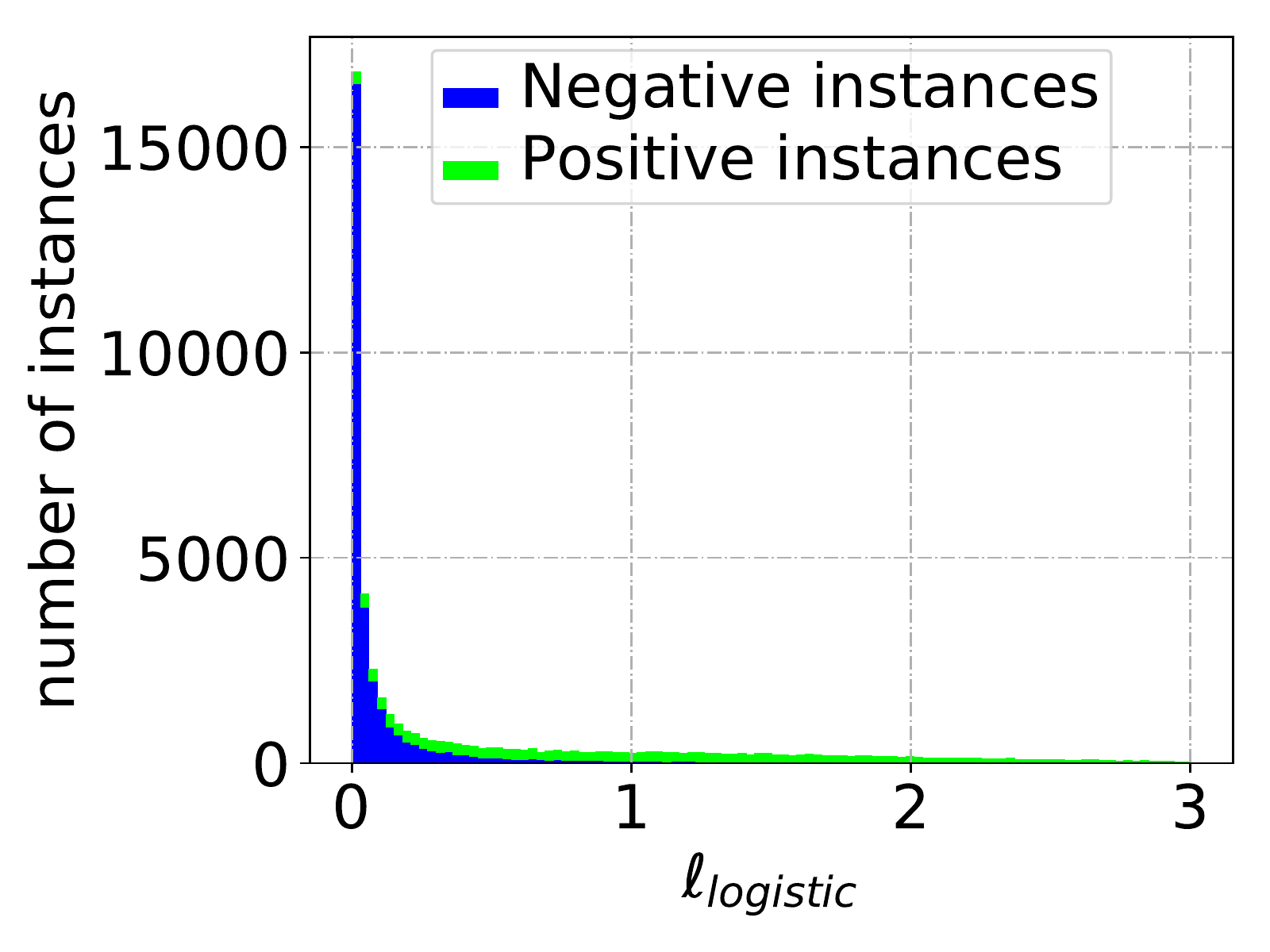}\\
\mbox{(a)} \small{Epoch $10$}
\end{minipage}
\begin{minipage}[h]{2.0in}
\centering
\includegraphics[width= 2.0in]{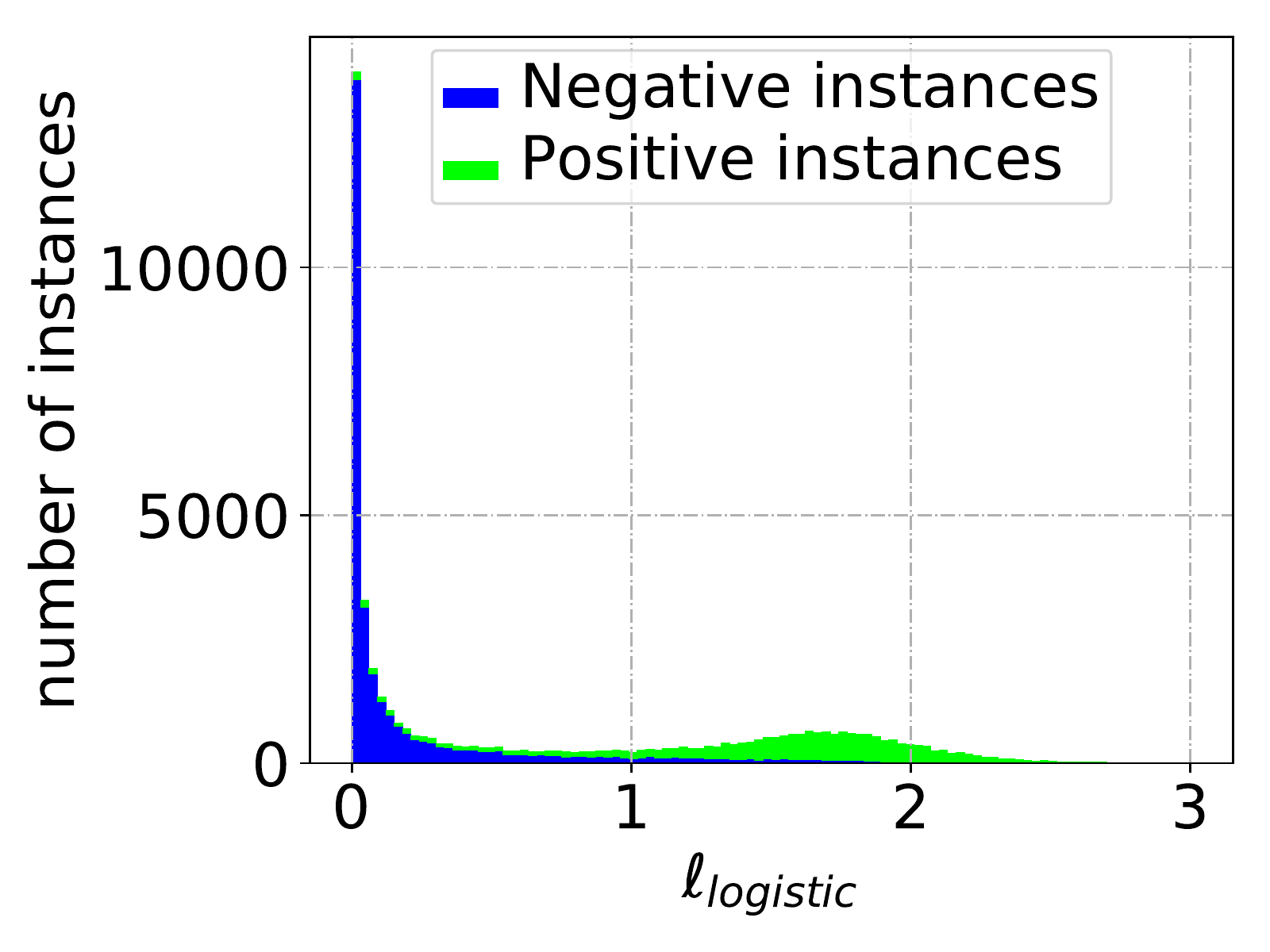}\\
\mbox{(b)} \small{Epoch $50$}
\end{minipage}
%\begin{minipage}[h]{2.0in}
%\centering
%\includegraphics[width= 2.0in]{img/fig3/hist_100}\\
%\mbox{(c)} Epoch $100$
%\end{minipage}
\begin{minipage}[h]{2.0in}
\centering
\includegraphics[width= 2.0in]{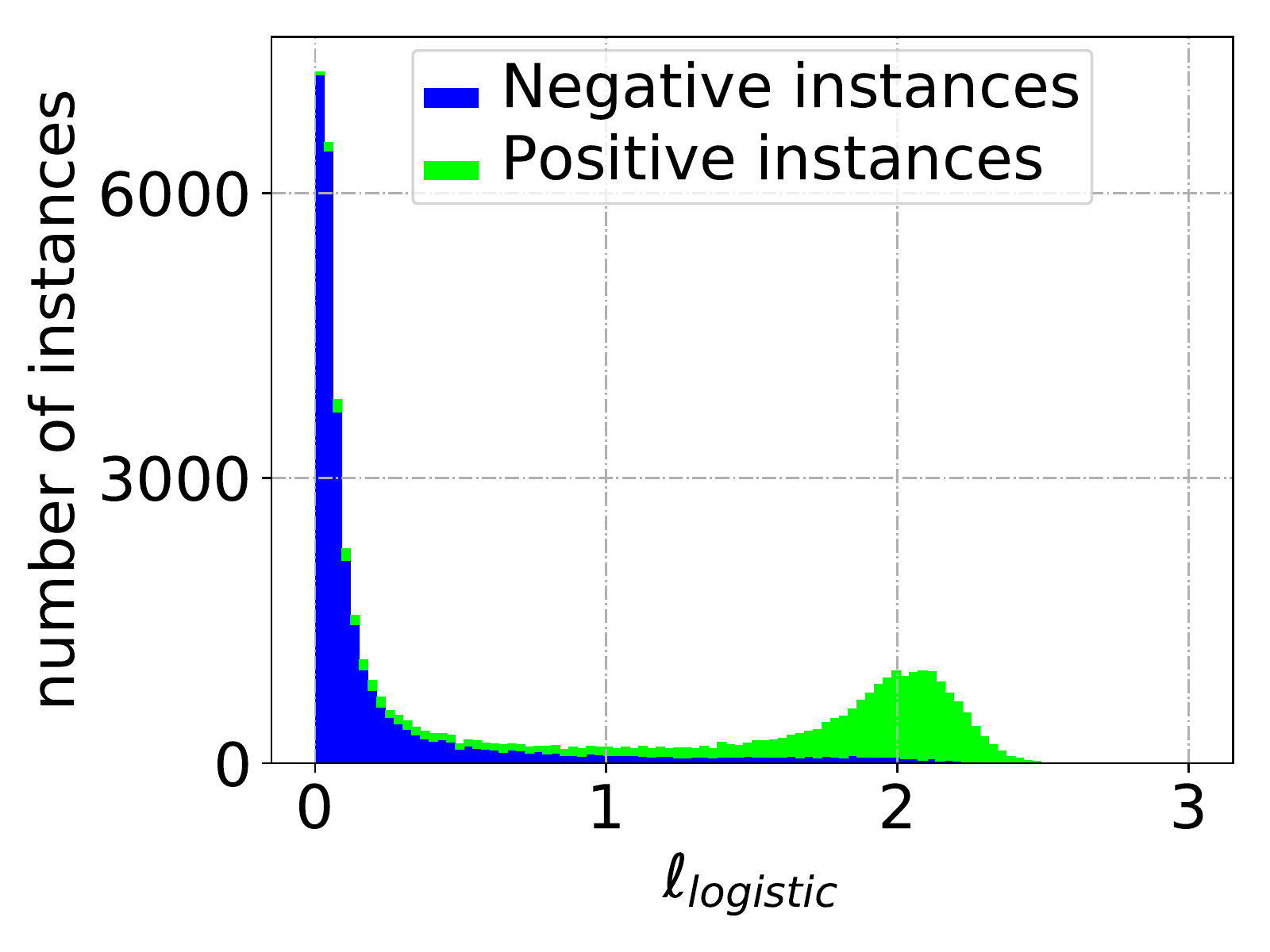}\\
\mbox{(c)} \small{Epoch $200$}
\end{minipage}
\caption{\small  The nnPU algorithm's logistic loss histogram of U data on CIFAR-10. We feed forward all U data into the neural network and calculate their sigmoid loss when treating them as negative. We then divide the range of all the U data's logistic losse qually into $100$ bins and calculate the number of instances falling into each bin. (a) Sigmoid loss histogram on epoch $10$; (b) sigmoid loss histogram on epoch $50$; (c) sigmoid loss histogram on epoch $200$.
}\label{fig:loss-distribution-logistic-nnpu}
\end{figure*}

We also plot the histogram of the logistic loss calculated on U data in epochs $10$, $50$, and $200$, as shown in Figure~\ref{fig:loss-distribution-logistic-nnpu}. Comparing Figures~\ref{fig:loss-distribution-nnpu} and~\ref{fig:loss-distribution-logistic-nnpu}, we can see that in an appropriate epoch, such as epoch $50$, there is a loss range in which we can select pure positive data, if setting the threshold appropriately. In this way, we will use the logistic loss as a surrogate loss to perform sample selection. 

\subsection{How to Use the Selected Data}
With the selected data, one trivial way to use them is to add them directly into the set of P data. However, we found such a simple operation cannot work well. In this section, we will discuss on this problem.

\begin{figure*}[!t]
\centering
%\begin{minipage}[h]{2in}
%\centering
%\includegraphics[width= 2in]{img/fig4/100}\\
%\mbox{(a) Add 100 instances}
%\end{minipage}
\begin{minipage}[h]{2in}
\centering
\includegraphics[width= 2in]{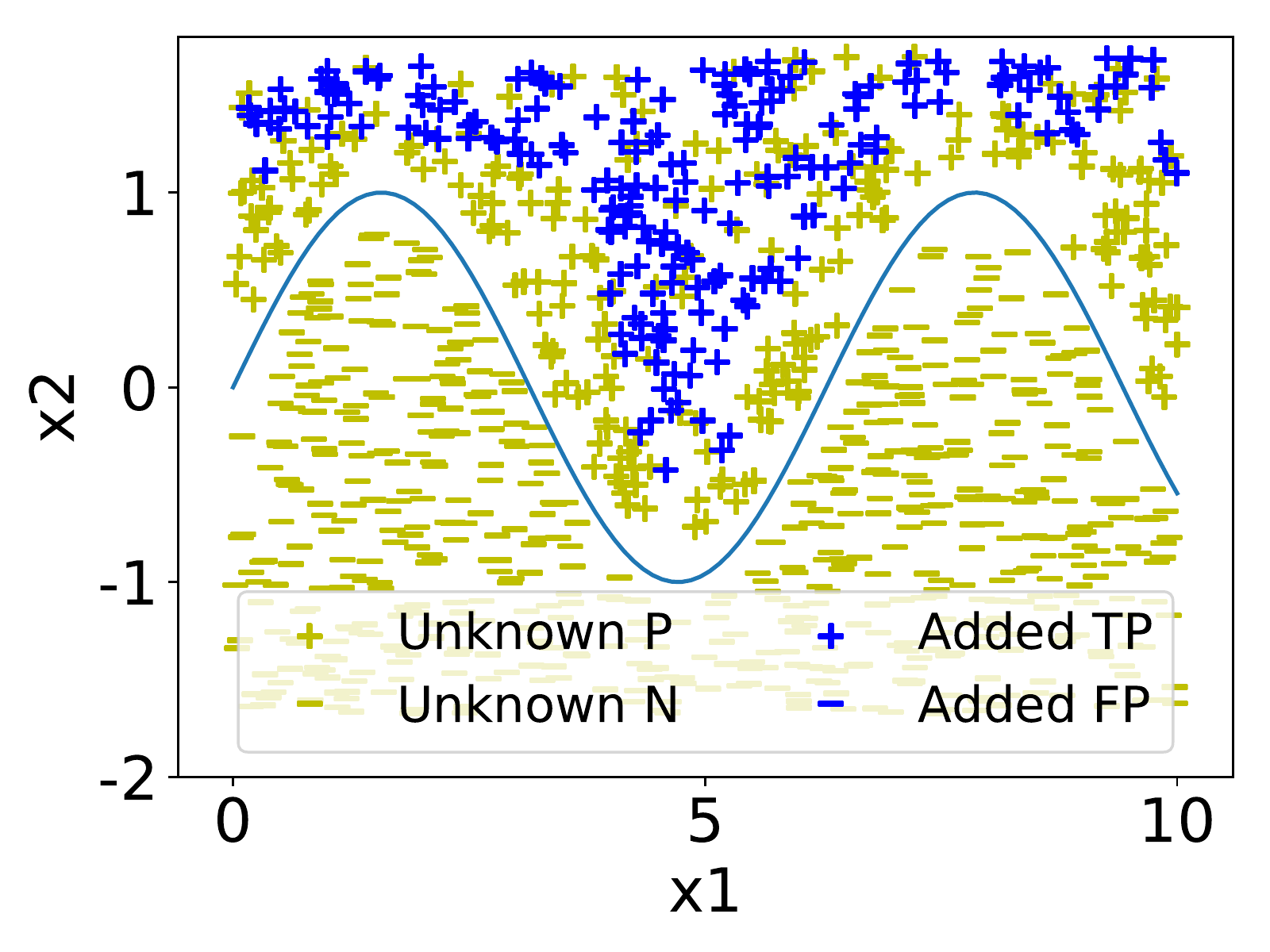}\\
\mbox{(a) \small{Top 200 large-loss U data}}
\end{minipage}
\centering
\begin{minipage}[h]{2in}
\centering
\includegraphics[width= 2in]{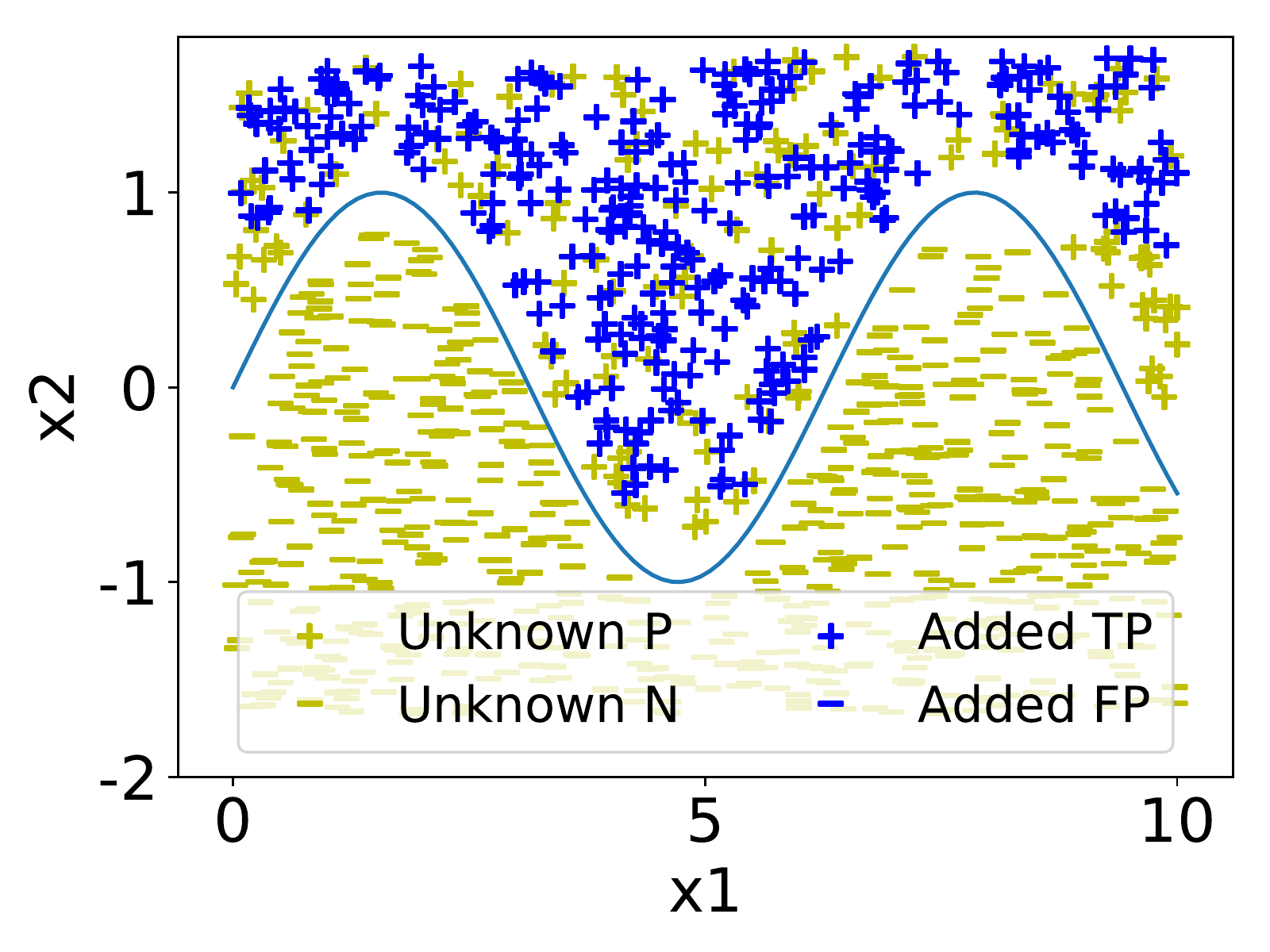}\\
\mbox{(b) Top 300 large-loss U data}
\end{minipage}
\begin{minipage}[h]{2in}
\centering
\includegraphics[width= 2in]{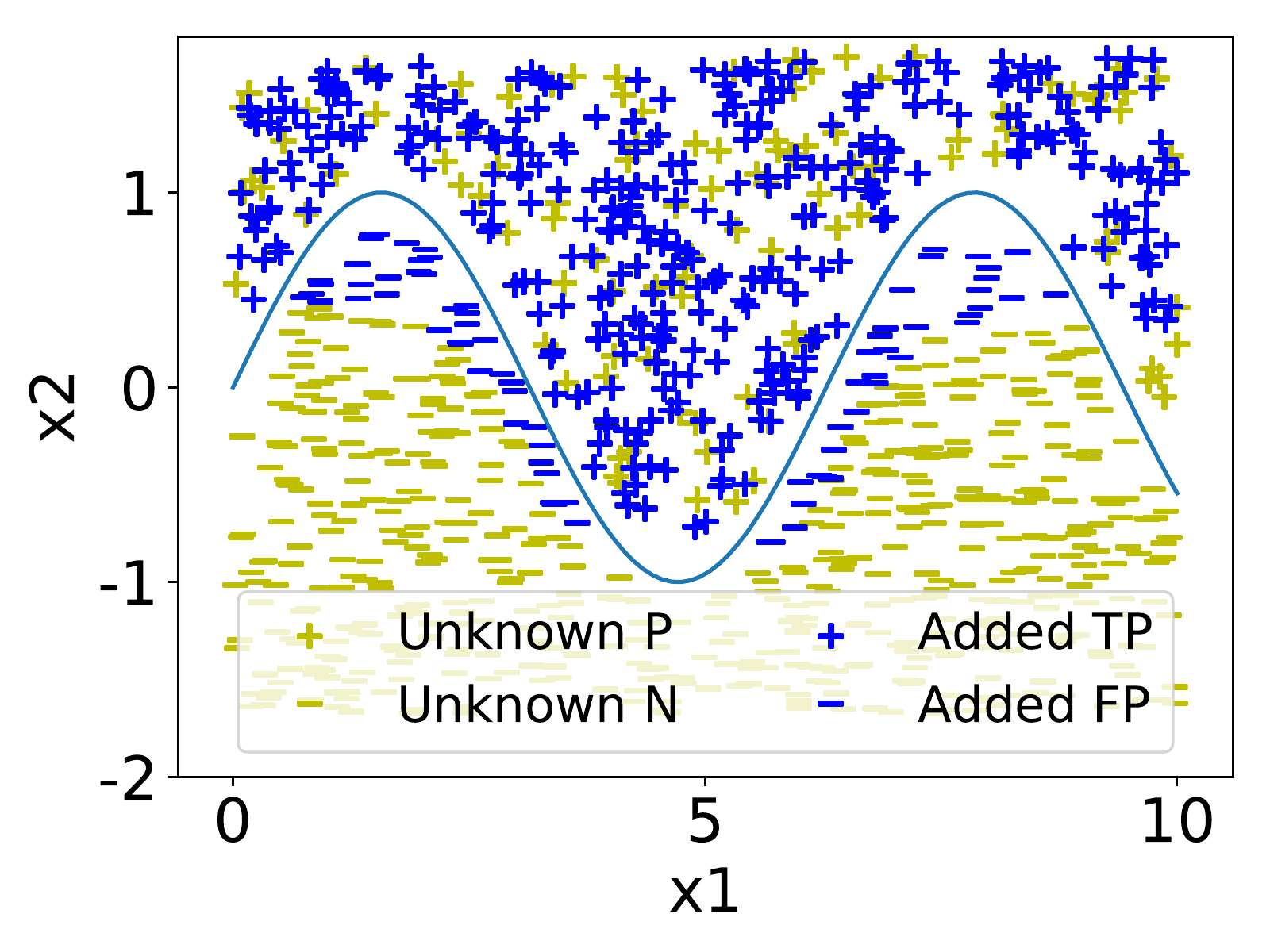}\\
\mbox{(c) Top 400 large-loss U data}
\end{minipage}
\caption{\small   Experimental results showing the data with largest loss vs. decision boundary in epoch $200$. (a) The top $200$ largest loss data. (b) The top $300$ largest loss data. (c) The top $400$ largest loss data.}\label{fig:positive-vs-boundary}
\end{figure*}

In Eq.~(\ref{eq:negative-loss}), one principle is to use the available P data to calculate their risk when they are treated as N, i.e.,
\begin{eqnarray}\label{eq:p-data-n-risk}
\widehat R_\mathrm{p}^-=\frac{1}{n_\mathrm{p}}\sum_{i=1}^{n_\mathrm{p}} \ell(-g(x_i;\theta)).
\end{eqnarray}
$\widehat R_\mathrm{p}^-$ is an unbiased estimation of $R_\mathrm{p}^-$ only when the P data are not biased and can represent the true distribution of $p(x|Y=1)$. However, we perform some experiments on synthetic data using the setting of Sec.~\ref{sec:synexp} and find the selected data cannot represent the true distribution of P data. More specially, we selected $100$ to $300$ large-loss data in epoch $200$, and plotted them against the decision boundary as they are 2-dimensional synthetic data. The results are shown in Figure~\ref{fig:positive-vs-boundary} (a) (b) (c). We can see that to keep the sampled data containing less negative data, we should sample a safe amount of them, otherwise if we sample $400$ large-loss data we can easily meet a lot of false positive. To keep the selected data clean, we need sample only a safe amount of large-loss data, which may not cover all the positive data space and have a large tendency to be \emph{biased}.

Note that in PU learning, we need to calculate two items: $R_\mathrm{p}^+$ and $R_\mathrm{n}=R_\mathrm{u}-\pi R_\mathrm{p}^-$ as we show in Eq.~(\ref{eq:expect-pu-risk}). There is almost no problem with calculating $R_\mathrm{p}^+$ by adding these small-margined P data. However, when calculating $R_\mathrm{n}$, the P data are always treated as negative ones, and the small-margined P data will result in serious problem. Specially, when the small-margined P data are treated as negative, their empirical surrogate losses, $\widehat R_\mathrm{p}^-$ in~Eq.~(\ref{eq:p-data-n-risk}) could be very large as they are in the positive area far away from the decision boundary. In this way, subtracting it in Eq.~(\ref{eq:empirical-pu-risk}) would lead to an underdestimation of the empirical risk which will diverge greatly compared to the unbiased risk under this situation. 

Target at this problem, in the following, we design a new learning objective. Given the P data $\mathcal{X}_\mathrm{p}$, U data $\mathcal{X}_\mathrm{u}$ and the selected positive instance $\mathcal{S}$, we have
\begin{eqnarray*}
\widehat R_\mathrm{aaPU}=\frac{\pi}{n_\mathrm{p}} \sum_{x_i\in \mathcal{X}_\mathrm{p}\cup \mathcal{S}} \ell_\mathrm{logistic}(g(x_i;\theta))\;\;\;\;\;\;\;\;\;\;\;\;\;\\
+\max\left(\frac{1}{n_\mathrm{u}}\sum_{x_i\in\mathcal{X}_\mathrm{u}} \ell_\mathrm{logistic}(-g(x_i;\theta))-\right.\;\;\;\;\;\;\;\;\\
\left.\frac{\pi}{n_\mathrm{p}}\sum_{x_i\in\mathcal{X}_\mathrm{p}} \ell_\mathrm{logistic}(-g(x_i;\theta)),0\right).
\end{eqnarray*}
Here the main difference between $\widehat R_\mathrm{aaPU}$ and $\widehat R_\mathrm{nnPU}$ is that, instead of treating the selected data simply as P data and use them everywhere in Eq.~(\ref{eq:loss-nnpu}), we only use them to estimate the \emph{positive} risk $R_\mathrm{p}^+$, but not use them to estimate the negative risk $R_\mathrm{n}$. Such a difference will help us to get rid of the impact of biasness in selected data.

\subsection{Algorithm}
To summarize, we will call our method \emph{adaptively augmented PU learning}  (aaPU).  For existing PU learning methods based on complex models, aaPU works as a wrapper to automatically identify P data from a set of U data during the training process. In each epoch, P data is selected from U data and will be used for future training. A framework of our proposed algorithm is shown in Algorithm~\ref{alg:aapu}. 

We first initialize the parameter $\theta$ and $\mathcal{S}$ (Line $1$), and in each epoch, we shuffle the current P data, U data and $\mathcal{S}$, dividing them into small batches (Line $3$ to $4$). Note that we keep the ratio between P and U data in each mini batch to be the same as the batch P and U data. We then use these mini-batches to do stochastic gradient descent (Line $5$ to $6$). After updating the network by all mini-batches in one epoch, we will feed forward all the U data into the current neural network and calculating their surrogate loss using $Y=-1$ as the ground truth (Line $7$). We thus get $L_u$, which is
\begin{eqnarray}\label{eq:lu}
L_u=[\ell(-g(x_1;\theta)),\ldots,\ell(-g(x_{n_\mathrm{u}};\theta))] 
\end{eqnarray}
for all $x_i\in\mathcal{X}_\mathrm{u}$ and $i=[n_\mathrm{u}]$. Finally we select the instances with the first $n_s$ largest surrogate losses and not in $\mathcal{X}_\mathrm{p}$ (Line $8$) and put these selected P data into $\mathcal{S}$ (Line $9$), which will be together with the original P data to estimate the positive risk $R_\mathrm{p}^+$. Through the procedure of aaPU, the positive data can be added adaptively and permanently within the evolving of training neural networks.

\begin{algorithm}[!t]
	\centering
	\caption{adaptively augmented PU learning (aaPU)}
	\label{alg:aapu}
	\begin{algorithmic}[0]
	%\small
	\Input
  \Desc{$\mathcal{X}_\mathrm{p}$}{positive training data}
  \Desc{$\mathcal{X}_\mathrm{u}$}{unlabeled training data}
  \Desc{$n_\mathrm{u}$}{number of unlabeled training data}%, $n_\mathrm{u}=|\mathcal{X}_\mathrm{u}|$}
  %\Desc{$m_\mathrm{p}$}{mini-batch size for positive data}
  %\Desc{$m_\mathrm{u}$}{mini-batch size for unlabeled data}
  \Desc{$T$}{maximum number of epochs}
  \Desc{$\eta$}{learning rate of stochastic gradient descent}
\Desc{$\mathcal{S}$}{selected data}
  \EndInput
  \Output
  \Desc{$\theta$}{model parameter $\theta$ for $g(x;\theta)$}
  \EndOutput 
  %\State Let $\mathcal{A}(\theta)=\argmin_\theta \widehat{R} (g;\mathcal{X}_\mathrm{p};\mathcal{X}_\mathrm{u})$
  %\State Let $\mathcal{A}(\theta)$ to gradient de
    \end{algorithmic}
	\begin{algorithmic}[1]
 \State \textbf{Initialize} $\theta$ and $\mathcal{S}=\emptyset$
  \For {$t=1,2,\ldots,T$}
  		\State Shuffle $\mathcal{X}_\mathrm{p},\mathcal{X}_\mathrm{u}$, $\mathcal{S}$ into $N_t$ mini-batches %with minibatch size $(m_\mathrm{p},m_\mathrm{u})$,
  		\State Denoted by $(\mathcal{X}_\mathrm{p}^i,\mathcal{X}_\mathrm{u}^i,\mathcal{S}^i)$ the $i$-th mini-batch
   		\For {$i=1,2,\dots,N_t$}
               %\State Update $\theta$ by $\mathcal{A}(\theta)=\argmin_\theta \widehat{R} (g;\mathcal{X}_\mathrm{p}^i;\mathcal{X}_\mathrm{u}^i)$
				\State Update $\theta$ by $\theta=\theta-\eta \Delta_\theta \widehat R_\mathrm{aaPU} (g;\mathcal{X}_\mathrm{p}^i;\mathcal{X}_\mathrm{u}^i;\mathcal{S}^i)$
		\EndFor
		%\State $L_u=[\ell(g(x_1;\theta),-1), \ell(g(x_2;\theta),-1),\ldots,\ell(g(x_{n_\mathrm{u}};\theta),-1)]$ for $x_i\in\mathcal{X}_\mathrm{u}$ and $i=[n_\mathrm{u}]$
            \State Calculate  $L_u$ using Eq.~(\ref{eq:lu})
        %\State $\mathcal{S}=\emptyset$
        %\State $\mathcal{S}_\mathrm{p}=\argmin_{x_i} L_{ui}$ s.t. $|\mathcal{S}_p|\le\gamma, x_i\notin \mathcal{X}_P$
		\State Select the first $n_s$ largest $L_{ui}$ keeping  $x_i\notin \mathcal{X}_\mathrm{p}$
             \State Update $\mathcal{S}$ using corresponding $x_i$-s
 		%\State $\mathcal{X}_\mathrm{p}=\mathcal{X}_\mathrm{p} \cup \mathcal{S}$
  \EndFor
	\end{algorithmic}
\end{algorithm}

\section{Experiments}\label{sec:experiments}

In this section, we give the experimental results of aaPU on both synthetic and real data. The code is implemented in Python and all the deep learning models are implemented with PyTorch~\footnote{https://pytorch.org/}. %All the experimens are run on [TBA].

\subsection{Experiments on Synthetic Data}\label{sec:synexp}
%Data Generate Process
We first run experiments on $2$-dimensional synthetic data. We randomly sample $x_1$ uniformly from $(0,10)$ and $x_2$ uniformly from $(-1.5,1.5)$. The decision boundary is $f(x_1)=sin(x_1)$. All $(x_1,x_2)$ with $f(x_1)<x_2$ are P; otherwise, they are N. We set $x_2=x_2+0.2$ if $y=1$, i.e., $x_2>sin(x_1)$; and set $x_2=x_2-0.2$ otherwise, to make it easy to visualize the decision boundary. In this way we construct a training set with $100$ P data and $1,000$ U data. We also construct a test set with $10,000$ instances, including $4,412$ P and $5,588$ N. The $\pi$ is thus set as $0.4412$. 

%Proposal's Parameter Setting (including Network Structure)
We then compare aaPU with the state-of-the-art deep PU method nnPU~\cite{DBLP:conf/nips/KiryoNPS17}, which has already shown to be superior to uPU~\cite{DBLP:conf/icml/PlessisNS15}. Since the decision boundary $f(x)=sin(x)$ cannot be approximated by simple linear model, we use a neural network with three layers: one input layer of $2$-dimension, a fully collected layer of $200$ neurons, another fully collected layer of $600$ neurons, and finally the output layer. Each layer uses ReLU~\cite{DBLP:journals/jmlr/GlorotBB11} as activation function and batch normalization~\cite{DBLP:conf/icml/IoffeS15}. We use logistic loss as the surrogate loss. The optimizer is Adam~\cite{DBLP:journals/corr/KingmaB14} with default parameters in PyTorch and we will not use any dropout~\cite{DBLP:journals/jmlr/SrivastavaHKSS14} for this simple task. The weight decay parameter is set as $0.05$. We first use $10^{-4}$ as learning rate, and lower it to $10^{-5}$ after the $100$th epoch. The batch size is set as $128$ and we run totally $1,000$ epochs. For aaPU, we begin to select positive data from the $200$th epoch. In each epoch one instance with the largest loss is added. For nnPU, we use the same hyper parameter setting as the original paper, i.e., $\beta=0$ and $\gamma=1$.

\begin{figure}[!tb]
\centering
\begin{minipage}[h]{2.6in}
\centering
\includegraphics[width= 2.6in]{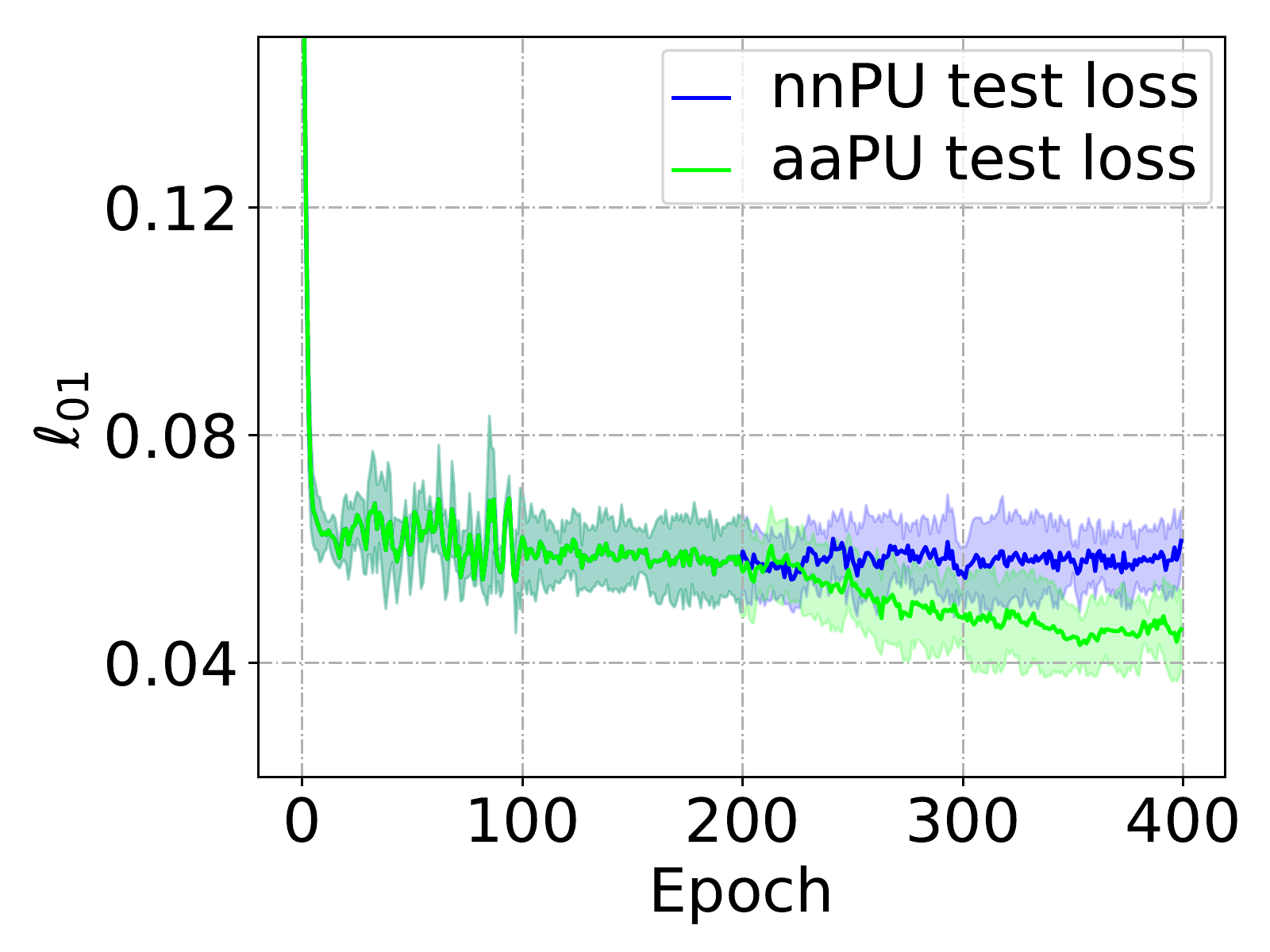}\\
%\mbox{(a)}
\end{minipage}
\caption{\small   Results on synthetic data comparing aaPU and nnPU, where the shadow shows the variance of each method. }\label{fig:simulation-result}
\end{figure}

%Results2
We show the performance comparing nnPU and the proposed aaPU in Figure~\ref{fig:simulation-result}. From the results, we can see that when we start to select positive data in epoch $200$ and use them from then on, the proposed aaPU achieves better result than nnPU quickly. As the learning continues and more data are selected, there exists an obvious gap between nnpu and aaPU. The performance of aaPU becomes better and better.  

\begin{table}[!t]
\caption{\small Statistical information on real data sets: \#p train (number of positive data during training), \#u train (number of unlabeled data during training), \#p test (number of positive data during testing) and \#n test (number of negative data during testing).}\label{tbl:real-statistical}
\begin{center}
%\scriptsize
%\begin{sc}
%\setlength\tabcolsep{8pt}
\begin{tabular}{ccccc}
\hline
Data set & \#p train& \#u train &\#p test &\#n test \\
%MNIST&&&&\\
CIFAR-10&$1000$&$50000$&$4000$&$6000$\\
CIFAR-100&$1000$&$50000$&$5000$&$5000$\\
20News&$1000$&$11314$&$3278$&$4254$\\
\hline
\end{tabular}
%\end{sc}
\end{center}
\end{table} 

\begin{figure*}[!htpb]
\centering
\begin{minipage}[h]{2in}
\centering
\includegraphics[width= 2in]{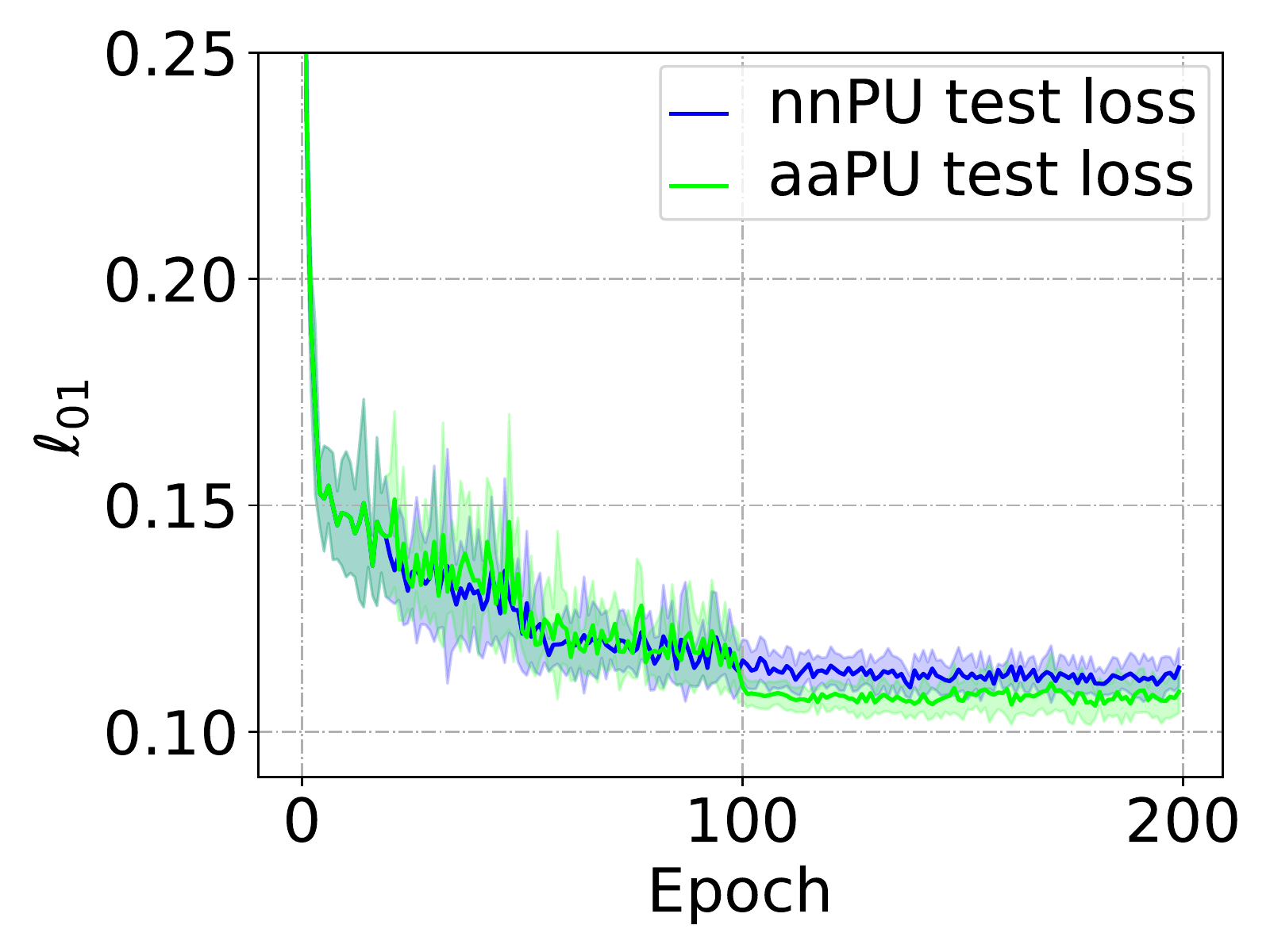}\\
\mbox\small{\small{CIFAR-10}}
\end{minipage}
\begin{minipage}[h]{2in}
\centering
\includegraphics[width= 2in]{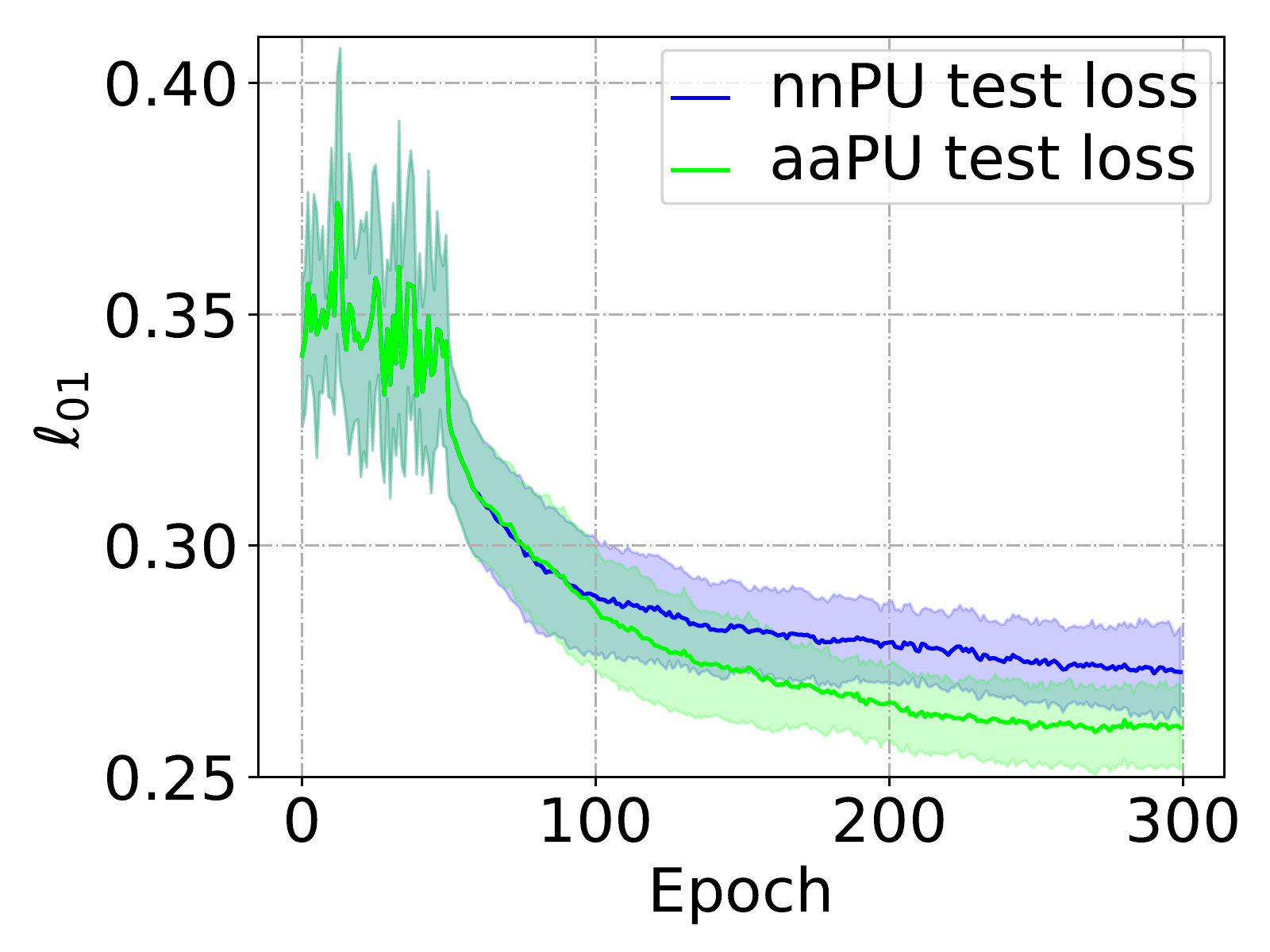}\\
\mbox\small{\small{CIFAR-100}}
\end{minipage}
\begin{minipage}[h]{2in}
\centering
\includegraphics[width= 2in]{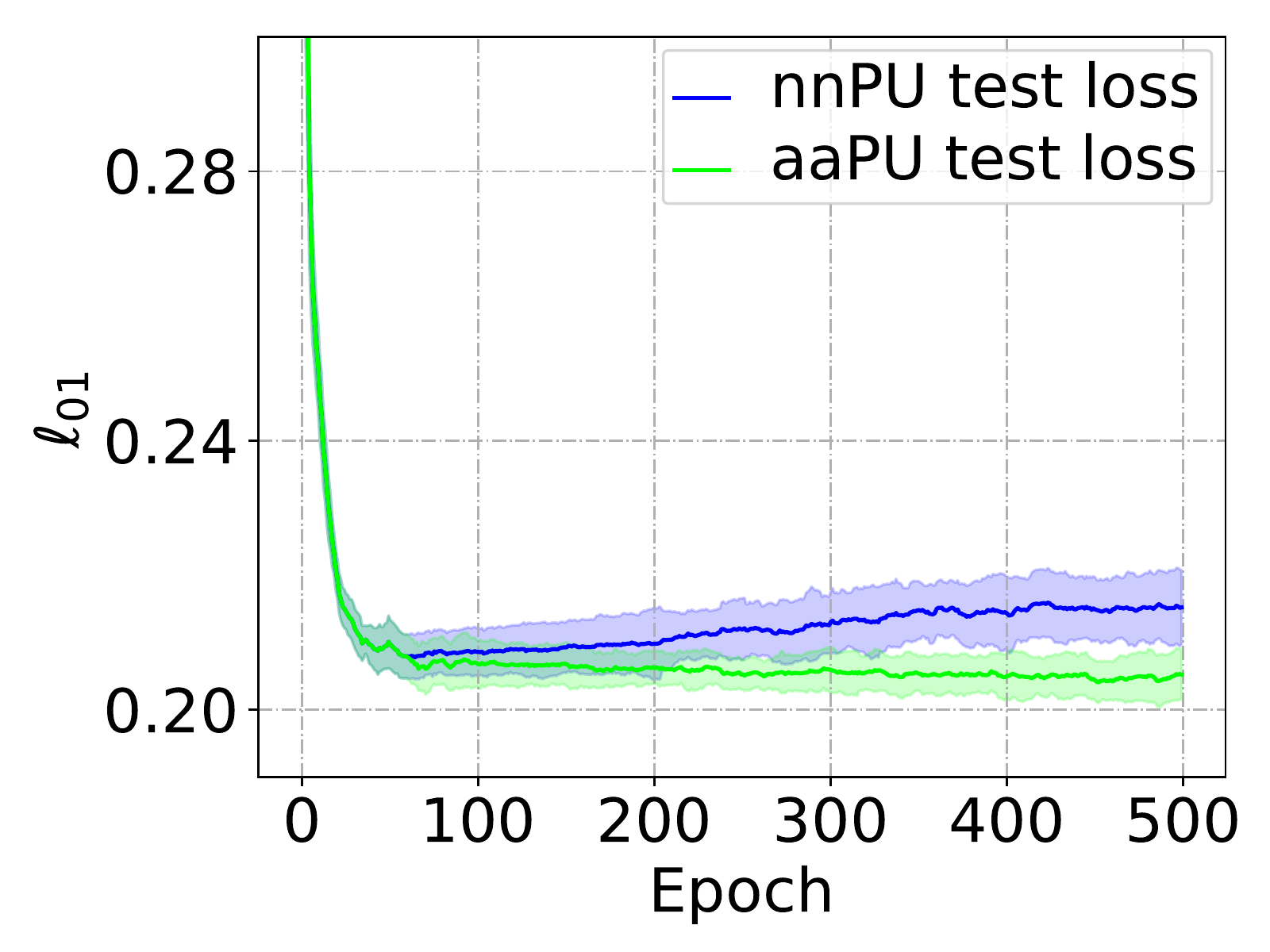}\\
\mbox\small{\small{20News}}
\end{minipage}
\caption{\small Test error on real data comparing aaPU and nnPU, where the shadow shows the variance of each method.}\label{fig:real-result}
\end{figure*} 

\subsection{Experiments on Real Data}\label{sec:exp-real}
In this section, we show the results on real data. We use three datasets: CIFAR-10~\cite{Krizhevsky09learningmultiple}, CIFAR-100~\cite{Krizhevsky09learningmultiple} and 20News~\cite{DBLP:conf/icml/Lang95}. CIFAR-10 is an image data set with three channels. CIFAR-100 is also an image data set similar to but with larger scale than CIFAR-10. 20News is a text data set. CIFAR-10 and 20News are used in previous PU works~\cite{DBLP:conf/nips/KiryoNPS17} and we use CIFAR-100 to show the scalable ability of aaPU.  

All these datasets are multi-class datasets. To make them suitable for binary classification, we adopt the same method as previous works~\cite{DBLP:conf/nips/KiryoNPS17} to make 20News into binary data. For CIFAR-100, we adopt the same method as CIFAR-10, dividing the data into animal/non animal\footnote{Positive: aquatic\_mammals, fish, insects, large\_carnivores, large\_omnivores\_and\_herbivores, medium\_mammals, non-insect\_invertebrates, people, reptiles, small\_mammals; 
Negative: flowers, food\_containers, fruit\_and\_vegetables, household\_electrical\_devices, household\_furniture, large\_man-made\_outdoor\_things, large\_natural\_outdoor\_scenes, trees, vehicles\_1, vehicles\_2}. For each data set, we randomly sample $\#p$ positive data according to $p(x|Y=1)$ to form P, and $\#u$ data according to $p(x)$ to form U. The detailed statistical information for each dataset can be found in Table~\ref{tbl:real-statistical}.

%Proposal's Parameter Setting (Including Network Structure)
On the three data sets, we use three different network structures as models for both aPU and nnPU, depending on the scale and type of data, similar to previous work~\cite{DBLP:conf/nips/KiryoNPS17,DBLP:conf/nips/HanYYNXHTS18}. 
%\vspace{-3mm}
\begin{itemize}%[leftmargin=*]
\item For CIFAR-10, we use the same network structure as~\cite{DBLP:conf/nips/KiryoNPS17}. We add dropout on the last three fully-connected layers with parameter $0.5$. The weight decay parameter is set as $0.017$. We use Adam as the SGD optimizer with $10^{-4}$ learning rate in the first $50$ epochs and $5\times 10^{-5}$ from epoch $50$ to $100$, and $10^{-5}$ after epoch $100$. Batch size is set as $2,000$, with totally $200$ epochs. For aaPU, from the $20$th epoch, $150$ instances are added in each epoch.
\item For CIFAR-100, we use the same network structure as~\cite{DBLP:conf/nips/HanYYNXHTS18}, with a dropout parameter as $0.5$. The weight decay parameter is set as $0.1$. We use Adam as the SGD optimizer with $10^{-4}$ learning rate in the first $50$ epochs and $10^{-6}$ from epoch $50$. Batch size is set as $1,000$, with totally $300$ epochs. For aaPU, from the $60$th epoch, $160$ instances are added in each epoch. 
\item For 20NewsGroup, we adopt the same network structure as~\cite{DBLP:conf/nips/KiryoNPS17} without dropout. The weight decay parameter is set as $5\times 10^{-4}$. We use Adam as the SGD optimizer with $10^{-4}$ learning rate in the first $50$ epochs and $10^{-5}$ after that. Batch size is set as $5,000$, with totally $500$ epochs. For aaPU, the weight decay and learning rate parameter is set the same as nnPU. We begin to add positive data from the $60$th epoch, and in each epoch we add $160$ instances. 
\end{itemize}
%\vspace{-3.5mm}
We repeat each experiments $10$ times and report the average results as well as the variance. The results are shown in Figure~\ref{fig:real-result}. We can see that, an all three datasets, our proposed aaPU get better results. Specially, aaPU is better on CIFAR-10, and much better on CIFAR-100 and 20News. The superior of aaPU begins to appear tens of epochs after selecting P data.

%\vspace{-2mm}
\paragraph{Discussion: }
In aaPU, one important factor is to decide when to add the selected data from the set of U data, and how much to add each time. For the first question, we usually begin to add P data when the validation loss becomes stable. %If added earlier, the classifier is not able to differentiate P and N data; if added later, the network has already memorized all the data, making it difficult to use ``large-small loss trick''. 
%In practice, we usually use a large learning rate from the beginning, and reduce it then. 
In practice, we find that the validation loss becomes stable usually $10$ epochs after reducing the learning rate. For the second question of how much to add, we cannot add too much data for that too much will bring more false positive. But if we add only a small amount of data, since such data are too far away from the decision boundary, they will have little contribution to train the classifier. In practice, 
we select the number of positive data selected from $[40,80,160,320]$. %Usually, the larger the number the better the performance. 

\section{Conclusion}\label{sec:conclusion}

In this paper, to answer whether there is a new sample selection method that can outperform the latest importance reweighting method in the deep learning age, we propose aaPU. aaPU automatically selects P data from U data during training, and uses these selected data in future epochs. Specially, aaPU uses deep networks as classification model. After each training epoch, all U data are fed forward into the neural networks, and logistic loss of them are calculated using $-1$ as ground truth. Then data with large loss are selected to estimate the positive risk, which is part of the risk minimized in further training. Experiments validate that the proposed aaPU can work better than nnPU~\cite{DBLP:conf/nips/KiryoNPS17} on both synthetic and real data. Thus sample selection approach is possible to outperform the latest importance reweighting approach in the deep learning age. 

In the future, we plan to combine aaPU with~\cite{DBLP:journals/corr/abs-1810-00846}, which studies the learning problem when P data, U data and biased N data are presented. Although~\cite{DBLP:journals/corr/abs-1810-00846} is not targeted at selecting data, the technology proposed there dealing with biased N data are complementary to our work, and may be used to further improve aaPU. 

\subsubsection*{Acknowledgments} We want to thank Yu-Guan Hsieh for kindful suggestions on the experiments. MS was supported by JST CREST JPMJCR1403.

\bibliography{ref}
\bibliographystyle{ieeetr}

\end{document}